\documentclass[sn-basic, iicol]{sn-jnl}

\usepackage{graphicx}%
\usepackage{multirow}%
\usepackage{amsmath,amssymb,amsfonts}%
\usepackage{amsthm}%
\usepackage{mathrsfs}%
\usepackage[title]{appendix}%
\usepackage{xcolor}%
\usepackage{textcomp}%
\usepackage{manyfoot}%
\usepackage{booktabs}%
\usepackage{listings}%

\usepackage{bbding}
\usepackage{multirow}
\usepackage{bm}
\usepackage{colortbl}
\usepackage{scalerel,amssymb}
\usepackage{wasysym}
\usepackage{yhmath}
\usepackage{amsmath}
\usepackage{soul}
\usepackage[linesnumbered,ruled,vlined]{algorithm2e} 
\usepackage{algpseudocode}  

\definecolor{_version1}{RGB}{230,185,181}
\definecolor{_version2}{RGB}{216,228,192}
\definecolor{_version3}{RGB}{183,222,232}
\newcommand{\version}[1]{$\begingroup\color{#1}\CIRCLE\endgroup$}

\definecolor{lightblue}{RGB}{51,153,255}
\definecolor{lightyellow}{RGB}{245,157,86}
\definecolor{lightgreen}{RGB}{157,187,97}

\begin{document}

\title[Article Title]{P2P: Part-to-Part Motion Cues Guide a Strong Tracking Framework for LiDAR Point Clouds}


\author[1]{\fnm{Jiahao} \sur{Nie}}

\author[2]{\fnm{Fei} \sur{Xie}}

\author[3]{\fnm{Sifan} \sur{Zhou}}

\author[4]{\fnm{Xueyi} \sur{Zhou}}

\author[4]{\fnm{Dong-Kyu} \sur{Chae$^\dag$}}

\author[1]{\fnm{Zhiwei} \sur{He$^\dag$}}

\affil[1]{\orgdiv{Hangzhou Dianzi University}, \city{Hangzhou}, \country{China}}

\affil[2]{\orgdiv{Shanghai Jiao Tong University}, \city{Shanghai}, \country{China}}

\affil[3]{\orgdiv{Carnegie Mellon University}, \city{Pittsburgh}, \country{USA}}

\affil[4]{\orgdiv{Hanyang University}, \city{Seoul}, \country{South Korea}}


\abstract{3D single object tracking (SOT) methods based on appearance matching has long suffered from insufficient appearance information incurred by incomplete, textureless and semantically deficient LiDAR point clouds. While motion paradigm exploits motion cues instead of appearance matching for tracking, it incurs complex multi-stage processing and segmentation module. In this paper, we first provide in-depth explorations on motion paradigm, which proves that (\textbf{i}) it is feasible to directly infer target relative motion from point clouds across consecutive frames; (\textbf{ii}) fine-grained information comparison between consecutive point clouds facilitates target motion modeling. We thereby propose to perform part-to-part motion modeling for consecutive point clouds and introduce a novel tracking framework, termed \textbf{P2P}. The novel framework fuses each corresponding part information between consecutive point clouds, effectively exploring detailed information changes and thus modeling accurate target-related motion cues. Following this framework, we present P2P-point and P2P-voxel models, incorporating implicit and explicit part-to-part motion modeling by point- and voxel-based representation, respectively. Without bells and whistles, P2P-voxel sets a new state-of-the-art performance ($\sim$\textbf{89\%}, \textbf{72\%} and \textbf{63\%} precision on KITTI, NuScenes and Waymo Open Dataset, respectively). Moreover, under the same point-based representation, P2P-point outperforms the previous motion tracker M$^2$Track by \textbf{3.3\%} and \textbf{6.7\%} on the KITTI and NuScenes, while running at a considerably high speed of \textbf{107 Fps} on a single RTX3090 GPU. The source code and pre-trained models are available at \url{https://github.com/haooozi/P2P}.}



\maketitle

\begingroup
\renewcommand\thefootnote{}\footnote{\textsuperscript{\dag}Corresponding authors}
\addtocounter{footnote}{-1}
\endgroup

\section{Introduction}
Single object tracking (SOT) is a long-standing topic in computer vision. Over the past years, tracking has mainly relied on 2D image data, witnessing many advanced techniques, such as Siamese region proposal network \citep{siamrpn,ttdimp,transt,dualtfr,supersbt} and one-stream framework \citep{mixformer,ost}. With the rapid development of LiDAR sensors recently, increasing efforts \citep{sc3d,p2b,m2track} have been devoted to 3D SOT on point clouds. Current methods~\citep{p2b,bat,cui20213d,shan2022real,cmtrack,hu2024mvctrack} mostly follow appearance matching paradigm that utilizes template target information to enhance target-specific feature of search region for subsequent target localization. Despite demonstrated success, incomplete and textureless point clouds introduce insufficient appearance information, limiting the application of appearance matching paradigm.

\begin{figure}[t]
\centering
\includegraphics[width=1\linewidth]{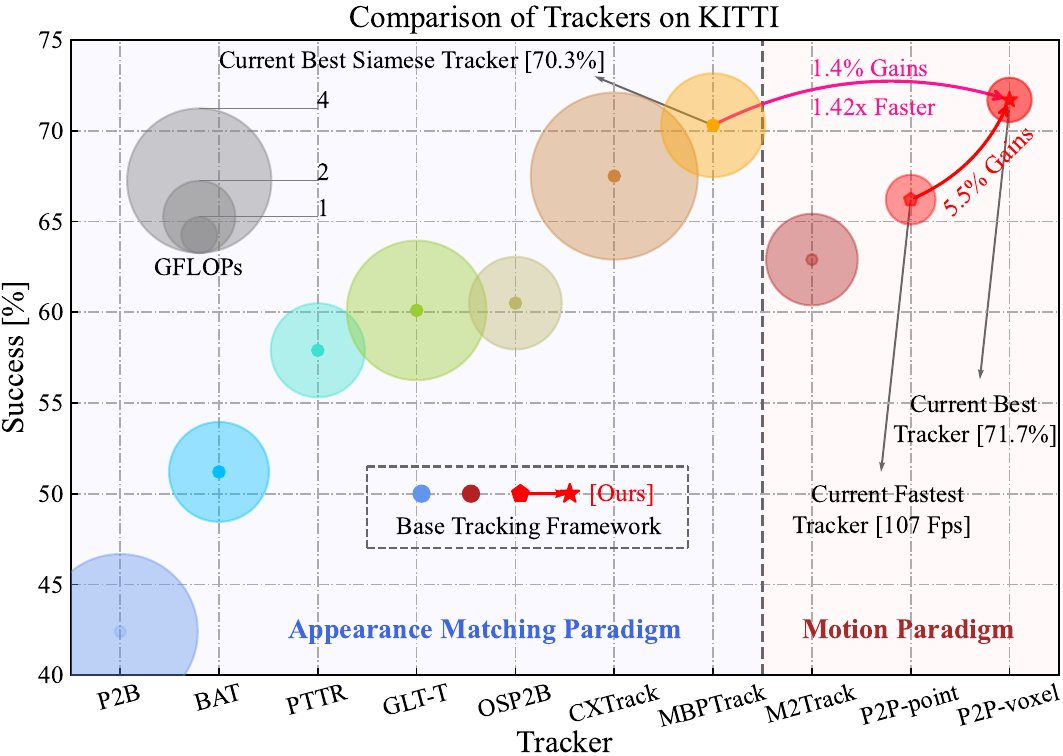}
\vspace{-10pt}
\caption{Comparison with state-of-the-art methods. We visualize mean success performance across all categories on KITTI dataset~\citep{kitti} with respect to floating-point operations per second (FLOPs). P2P-point and P2P-voxel indicate the proposed tracking models with point-based representation and voxel-based representation, respectively.}
\label{fig1}
\end{figure}

M$^2$Track~\citep{m2track} introduces a motion paradigm specialized for 3D SOT, achieving significant improvements in both performance and efficiency over appearance matching baseline P2B~\citep{p2b}, as shown in Fig.~\ref{fig1}. It first segments foreground points corresponding to tracked target from the cropped point clouds of two consecutive frames, and then infers 4-DOF\footnote{The length, width and height of target bounding box are given in the first frame, and are unchanged across all frames in a point cloud sequence.} relative motion of the target between the two frames. The relative motion is applied to the target position in previous frame to locate the target in current frame. Through a comprehensive analysis, we find that the effective utilization of background points can assist in target localization, which has also been explored and validated in CXTrack~\citep{cxtrack}. In addition, M-Vanilla~\citep{m2track++} further demonstrates the potential of the motion paradigm by removing additional auxiliary modules. Therefore, a question is naturally raised: \textit{Is it possible to infer target's 4-DOF relative motion directly from cropped point clouds of two consecutive frames?}

\begin{figure*}[t]
  \centering
   \includegraphics[width=1.0\linewidth]{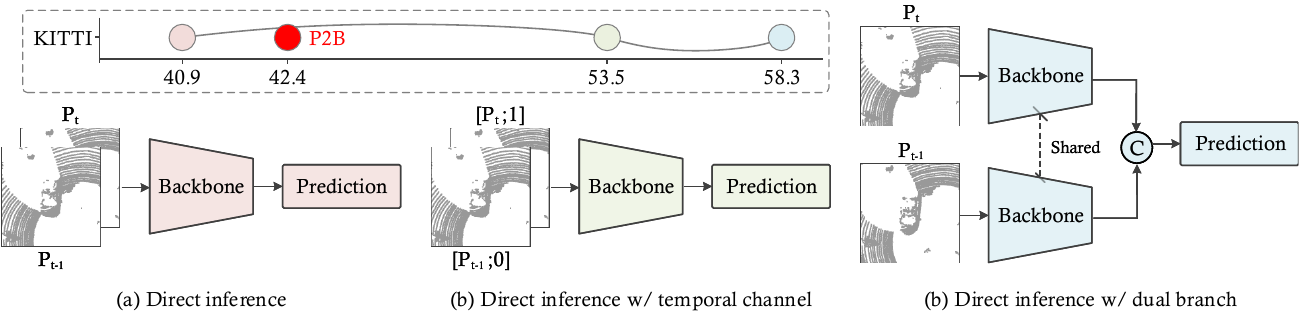}
   \vspace{-10pt}
   \caption{Preliminary investigation on KITTI dataset~\citep{kitti}. We conduct a series of experiments of (a), (b) and (c) that denoted by circles with different colors [\version{_version1},\version{_version2},\version{_version3}].}
   \label{fig2}
\end{figure*}

To investigate this question, we conduct a series of exploratory experiments, as shown in Fig.~\ref{fig2}. We first merge the cropped point clouds of two consecutive frames to directly infer the relative motion [\version{_version1}], using merely a regression loss. Despite simplicity, such a structure yields a slightly lower performance than P2B (40.9\% v.s. 42.4\%), which gives a positive answer to this question. Based on this structure, we introduce an additional temporal channel [\version{_version2}] to distinguish the previous and current frames. As expected, tracking performance improves by 12.6\%. Incorporating box-aware information leads to an additional performance boost, as demonstrated in M-Vanilla~\citep{m2track++}. We further explore an alternative solution [\version{_version3}] to distinguish the consecutive frames. It first independently extracts point cloud features from each frame via dual branches, and then concatenates them for prediction. To our surprise, the tracking performance is up to 58.3\%. One intuitive reason is that such an approach not only distinguishes between the previous and current frames, but also facilitates to compare the features between the two frames, learning fine-grained information changes and thereby enhancing motion cues.

Motivated by these exploratory experiments, we cast tracking task as a direct inference of target relative motion. To form accurate and robust motion cues, we propose to learn detailed information changes between each corresponding portion of consecutive point clouds, which we define as part-to-part motion modeling. We thereby introduce a novel tracking framework, termed \textbf{P2P} that effectively models target-related motion cues in each part of two-frame point cloud flow. Considering the disordered nature of LiDAR point clouds, a direct part-to-part fusion between the consecutive point clouds at the input level seems to be infeasible. Towards this problem, we leverage point and voxel representations to form ordered point cloud features with implicit and explicit spatial structure information, respectively. Therefore, we propose P2P-point and P2P-voxel, which fuse each corresponding part between the spatial structure of consecutive point clouds in feature-level. Benefiting from part-to-part fusion, fine-grained motion cues can be generated, thereby guiding accurate tracking for point cloud objects. Compared with M$^2$Track~\citep{m2track}, both of P2P-point and P2P-voxel get rids of auxiliary modules such as segmentation and two-stage motion refinement, while achieving better performance and higher speed. Comprehensive experiments are conducted on KITTI~\citep{kitti}, NuScenes~\citep{Nuscenes} and Waymo Open Dataset (WOD)~\citep{waymo}. The results confirm that our proposed P2P is a strong tracking framework that achieves leading performance while maintaining high efficiency.

In summary, the contributions of this work can be outlined as follows:
\begin{itemize}
    \item We provide a series of exploration and analysis on the motion paradigm and propose a novel tracking framework, termed P2P, providing a new perspective on the design of tracking models. 
    \item We introduce P2P-point and P2P-voxel models, which explore fine-grained motion cues by incorporating implicit and explicit part-to-part motion modeling via point- and voxel-based representation, respectively. 
    \item We conduct comprehensive experiments on KITTI, NuScenes and WOD, demonstrating the leading performance and potential of our proposed framework. 
\end{itemize}

\section{Related Work}
\subsection{2D Single Object Tracking}
In the context of 2D SOT, it often makes sense to decompose the tracking problem into classification and regression sub-tasks~\citep{zhou2023fastpillars}. The former focuses on localizing the target center, while the latter is dedicated to estimating the target size. Early correlation filter based tracking methods, such as KCF~\citep{kcf} and ECO~\citep{eco}, face challenges in performing both of the two sub-tasks simultaneously. Siamese network based appearance matching methods have proven effective in addressing these challenges, thus establishing a solid foundation for advancements in this field. Prominent trackers include SiamFC~\citep{siamfc}, SiamRPN~\citep{3dsiamrpn}, SiamCAR~\citep{siamcar}, STARK~\citep{stark}, MixFormer~\citep{mixformer}, and other notable variants~\citep{ocean,videotrack,tmt,sbt,grm,diffusiontrack,samn,siamla}. Recently, appearance matching techniques have inspired many excellent works on 3D single object tracking on point clouds.

\subsection{3D Single Object Tracking}
The pioneering work within the context of 3D SOT is SC3D~\citep{sc3d}. It generates a set of candidate 3D bounding boxes using a Kalman filter, and selects one box with the highest similarity to the given template target as predicted result. However, SC3D is not an end-to-end framework and fails to run in real-time due to the exhaustive candidates. Next work, P2B~\citep{p2b} presents a 3D region proposal network leveraging VoteNet~\citep{votenet}. It significantly enhances tracking performance while enabling a real-time speed. Based on this strong baseline, many follow-ups~\citep{ptt,mlvsnet,bat,pttr,osp2b,glt,cxtrack,mbptrack,synctrack,wu20253d,nie2024towards} have emerged. Taking recent methods as examples, CXTrack~\citep{cxtrack} designs a target-centric transformer network to explore contextual information, and MBPTrack~\citep{mbptrack} improve CXTrack with a memory network and a box-prior localization network. Following recently popular one-stream tracking framework~\citep{mixformer,ost}, SyncTrack~\citep{synctrack} synchronizes feature extraction and matching.

While these methods have demonstrated impressive performance, point clouds are usually incomplete and lack textured structure, failing to provide sufficient appearance information for matching. Against this background, M$^2$Track~\citep{m2track} introduces a motion paradigm for tracking. It models the target motion between consecutive frames to infer the target position, yielding commendable tracking performance and speed. In this paper, we deeply investigate the motion tracking paradigm and propose P2P, a strong tracking framework for 3D SOT on LiDAR point clouds.

\begin{figure*}[t]
  \centering
   \includegraphics[width=1.0\linewidth]{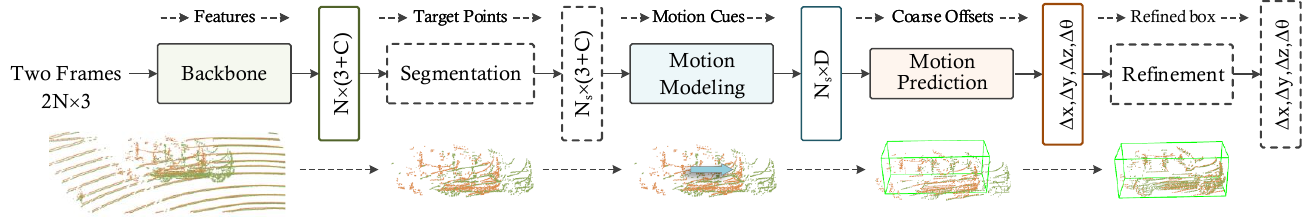}
   \vspace{-10pt}
   \caption{Framework of the existing motion paradigm~\citep{m2track}. M$^2$Track is the first motion tracker for 3D single object tracking with auxiliary components, such as segmentation module and motion refinement module.}
   \vspace{-10pt}
   \label{fig3}
\end{figure*}

\begin{figure*}[t]
   \includegraphics[width=0.825\linewidth]{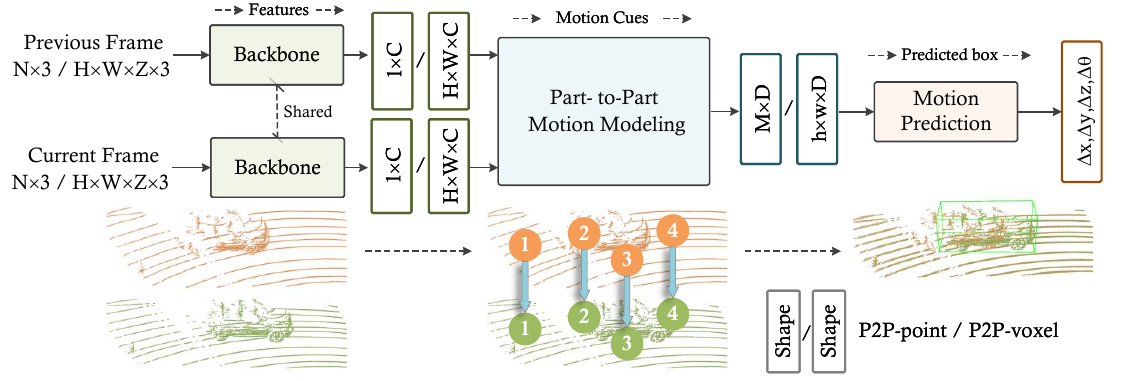}
   \caption{Framework of the proposed \textbf{P2P}. P2P is an end-to-end framework for 3D single object tracking without any auxiliary components. It is only composed of a feature extractor, a part-to-part motion modeling module and a motion prediction head.}
   \label{fig4}
\end{figure*}

\section{Methodology}

\subsection{Overview}
In this section, we present our tracking framework P2P, along with two tracking versions, \textit{i.e.}, P2P-point and P2P-voxel. To provide a better understanding, we first introduce the definition of 3D single object tracking task in Sec.~\ref{sec3.1}, followed by a thorough revisit of the existing motion tracking paradigm in Sec.~\ref{sec3.2}. Then, we detail the proposed P2P framework and the corresponding tracking models using both point- and voxel-based representations in Secs.~\ref{sec3.3},~\ref{sec3.4} and~\ref{sec3.5}, respectively.

\subsection{Task Definition}
\label{sec3.1}
The 3D single object tracking task on LiDAR point clouds is defined as: given a sequence of point clouds $\{\mathcal{P}_t\}_{t=1}^{T}$ consisting of $T$ frames and an initial target bounding box $\mathcal{B}_1=(x_1,y_1,z_1,w_1,h_1,l_1,\theta_1)$ in the first frame, tracking aims to localize a series of boxes $\{\mathcal{B}_t=(x_t,y_t,z_t,w_t,h_t,l_t,\theta_t)\}_{t=2}^{T}$ corresponding to this given target in subsequent frames. Here, $(x,y,z)$ and $(w,h,l)$ represent the center coordinate and size, respectively, while $\theta$ represents the rotation angle around $up$-axis. In the tracking configuration, the target box size \textit{i.e.}, $(w_t,h_t,l_t)$ is unchanged across all frames, only the parameters $(x_t,y_t,z_t,\theta_t)\in\mathbb{R}^4$ need to be predicted for $\mathcal{B}_t$.

\subsection{Revisit Motion Paradigm}
\label{sec3.2}
To predict $(x_t,y_t,z_t,\theta_t)$ of $\mathcal{B}_t$ for tracking, existing motion tracking paradigm~\citep{m2track} localizes the target by predicting its relative motion $(\Delta x_t,\Delta y_t,\Delta z_t,\Delta \theta_t)$ from previous frame $t-1$ to current frame $t$. As shown in Fig.~\ref{fig3}, this approach first generates search regions $\mathcal{P}_{t-1}^{crop}=\{p_{t-1}^i\}_{i=1}^N$ and $\mathcal{P}_{t}^{crop}=\{p_{t}^i\}_{i=1}^N$ ($N$ is the number of sampled points) for the two frames, centered on the previous prediction result $\mathcal{B}_{t-1}$. A spatial-temporal learning technique is then employed to segment foreground points in $\mathcal{P}_{t-1}^{crop}$ and $\mathcal{P}_{t}^{crop}$. The final relative motion of the target between two frames is inferred through a point set processing network. We can formulate this process as:
\begin{equation}
  [\Delta x_t,\Delta y_t,\Delta z_t,\Delta \theta_t] = F_{r}(F_{m}(F_{s}(\mathcal{P}_{t-1}^{crop},\mathcal{P}_{t}^{crop},\mathcal{B}_{t-1}))),
    \label{eq1}
\end{equation}
where $F_{s}$, $F_m$ and $F_{r}$ denote the segmentation module, motion inference module and motion refine module, respectively.

\subsection{Proposed P2P Framework}
\label{sec3.3}
The proposed P2P framework directly infers relative motion from $\mathcal{P}_{t-1}^{crop}$ and $\mathcal{P}_{t}^{crop}$, \textit{i.e.}, removing $F_{s}$ and $F_r$ in Eq.~\ref{eq1}, and embeds part-to-part motion modeling into $F_{m}$ to enhance target motion cues:
\begin{equation}
   [\Delta x_t,\Delta y_t,\Delta z_t,\Delta \theta_t] = F_{m}^{pp}(\mathcal{P}_{t-1}^{crop},\mathcal{P}_{t}^{crop},\mathcal{B}_{t-1}),
    \label{eq2}
\end{equation}
where $F_{m}^{pp}$ can mathematically express our P2P framework. Its overall architecture is illustrated in Fig.~\ref{fig4}. The framework consists of three components: feature extractor, part-to-part motion modeling, and motion prediction head. The feature extractor distinguishes and extracts spatial structure features of $\mathcal{P}_{t-1}^{crop}$ and $\mathcal{P}_{t}^{crop}$ by a weight-shared network using dual branches. The part-to-part motion modeling module first fuses each corresponding part between the spatial structures of $\mathcal{P}_{t-1}^{crop}$ and $\mathcal{P}_{t}^{crop}$, and then models target-related motion cues in each part. Finally, the prediction head infers the 4-DOF relative motion.

\subsection{Version \uppercase\expandafter{\romannumeral1}: P2P-point}
\label{sec3.4}

Here, we present the component details of our tracking version \uppercase\expandafter{\romannumeral1}: P2P-point, and describe its training and inference phases.

\noindent\textbf{Point Embedding.} Given a pair of point clouds, $\mathcal{P}_{t-1}^{crop}$ and $\mathcal{P}_{t}^{crop}$, cropped from frame $t-1$ and $t$ as inputs, we employ PointNet~\citep{pointnet}, a point-based representation network as the weight-shared feature extractor. More concretely, a series of weight-shared linear layers are used to extract features of $\mathcal{P}_{t-1/t}^{crop}\in\mathbb{R}^{N\times3}$, which is mapped into $f_{t-1/t}\in\mathbb{R}^{N\times C}$. Then, a global max-pooling operation is applied to maintain point permutation invariance, effectively capturing the spatial structure information of point clouds. The semantic features are further mapped into $\mathcal{F}_{t-1/t}\in\mathbb{R}^{1\times C}$.

\begin{figure*}[t]
  \centering
   \includegraphics[width=1.0\linewidth]{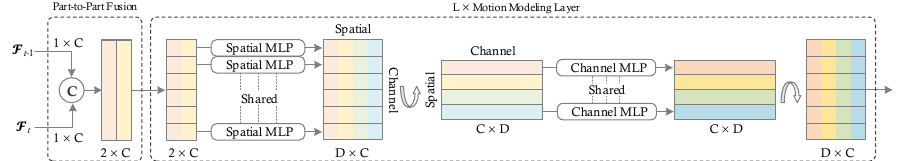}
   \vspace{-10pt}
   \caption{Illustration of part-to-part motion modeling module. It consists of a part to-part fusion operation and $L$ motion modeling layers, with each layer dedicated to feature learning at both spatial and channel levels.}
   \label{fig5}
\end{figure*}

\noindent\textbf{Part-to-Part Motion Modeling.} We introduce a part-to-part motion modeling module to model feature relationships to generate motion cues. As illustrated in Fig.~\ref{fig5}, it first concatenates $\mathcal{F}_{t-1}$ and $\mathcal{F}_{t}$ in the spatial dimension:
\begin{equation}
   \mathcal{F}^{pp}= {\rm Concat}([\mathcal{F}_{t-1}; \mathcal{F}_{t}], {\rm dim}=0),
    \label{eq3}
\end{equation}
where the output $\mathcal{F}^{pp}\in\mathbb{R}^{2\times C}$, which fuses each corresponding part between implicit spatial structures (as shown in Fig.~\ref{fig6}) of input two-frame point clouds. After that, we iteratively model motion cues of the two point clouds in both spatial and channel dimensions through cascaded 1D convolution layers~\citep{mlpmixer}, which can be mathematically expressed as:
\begin{equation}
   \mathcal{F}_{fusion}^{pp}=[{\rm Conv1D}([{\rm Conv1D}(\mathcal{F}^{pp})]^{\top})]^{\top},
    \label{eq4}
\end{equation}
where $\top$ denotes permutation operator. By this way, information changes at each part are captured, modeling more accurate and robust motion cues for motion prediction. Moreover, the resulting feature $\mathcal{F}_{fusion}^{pp}\in\mathbb{R}^{D\times C}$ enables P2P-point to distinguish distractors due to inconsistency of information changes in different parts.

\noindent\textbf{Prediction Head.} In contrast to the motion tracker M$^2$Track~\citep{m2track} that predicts relative motion while introducing a motion classifier and a refined previous target $\tilde{\mathcal{B}_{t-1}}$, we only apply a max-pooling operation along spatial dimension and a multi-layer perceptron (MLP) on the motion cue feature $\mathcal{F}_{fusion}^{pp}\in\mathbb{R}^{D\times C}$ to infer 4-DOF relative motion of the target:
\begin{equation}
   \mathcal{M}_{t-1,t} = {\rm MLP}({\rm Maxpooling}(\mathcal{F}_{fusion}^{pp})),
    \label{eq5}
\end{equation}
where $\mathcal{M}_{t-1,t}=(\Delta x_t,\Delta y_t,\Delta z_t,\Delta \theta_t)$ is used to transform the bounding box $\mathcal{B}_{t-1}$ to obtain $\mathcal{B}_{t}$.

\noindent\textbf{Training.} The proposed P2P-point can be trained in an end-to-end manner, using only a single regression loss function. To optimize the model to capture diverse motion distributions, we adopt residual log-likelihood estimation (RLE) loss~\citep{rle} as our regression loss.

\noindent\textbf{Inference.} During the inference phase, P2P-point predicts a series of parameters $\{\mathcal{M}_{t-1,t}\}_{t=1}^{T}$ for each frame in a point cloud sequence. Using these predicted parameters, we can perform a rigid body transformation $F_{trans}$ to update the previous prediction result $\mathcal{B}_{t-1} $ to obtain the current prediction target position $\mathcal{B}_t$:
\begin{equation}
   \mathcal{B}_{t} = F_{trans}(\mathcal{B}_{t-1}, \mathcal{M}_{t-1,t}).
    \label{eq6}
\end{equation}

\subsection{Version \uppercase\expandafter{\romannumeral2}: P2P-voxel}
\label{sec3.5}

Based on the proposed P2P framework, we further introduce tracking version \uppercase\expandafter{\romannumeral2}: P2P-voxel. While it shares foundational similarities with P2P-point in overall architecture, some notable differences are presented in terms of feature extractor, part-to-part motion modeling module and prediction head, as described below. More details could be found in Tab. \ref{table1}.

\noindent\textbf{Voxel Embedding.} To form voxel representations for the given point clods, we first partition the point clouds into 3D voxels $\mathcal{V}^{crop}_{t-1/t}\in\mathbb{R}^{W\times L\times H}$ using a grid-based approach~\citep{voxelnet}, and then employs a three-stage sparse convolution blocks as the feature extractor. The inputs $\mathcal{V}^{crop}_{t-1/t}\in\mathbb{R}^{W\times L\times H}$ are mapped into $\mathcal{F}_{t-1/t}\in\mathbb{R}^{\frac{W}{8}\times\frac{L}{8}\times D}$, where the height information is fused to the $D$-dimensional channel feature. Since voxels naturally have a spatial structure, we may avoid using pooling operators to ensure point permutation invariance to form implicit spatial structure. 

\noindent\textbf{Part-to-Part Motion Modeling.} Benefiting from voxel embedding that inherently preserves the spatial structure of point clouds, we could fuse each corresponding part between the explicit spatial structures of consecutive point clouds by directly concatenating the outputs of feature extractor $\mathcal{F}_{t-1}$ and $\mathcal{F}_{t}\in\mathbb{R}^{\frac{W}{8}\times \frac{L}{8}\times D}$ along the channels:
\begin{equation}
   \mathcal{F}^{pp}= {\rm Concat}([\mathcal{F}_{t-1}; \mathcal{F}_{t}], {\rm dim}=2),
    \label{eq7}
\end{equation}
where the obtained features $\mathcal{F}^{pp} \in\mathbb{R}^{\frac{W}{8}\times \frac{L}{8}\times 2D}$. Afterward, some simple 2D convolution blocks are used for target motion modeling, as illustrated in Tab.~\ref{table1}. By leveraging 2D convolution, both information changes in each corresponding spatial part and change inconsistency between different spatial parts are encoded, eliminating the need for the complex motion modeling structure adopted in P2P-point.

\noindent\textbf{Prediction Head.} Different from the max-pooling operation used in P2P-point for down-sampling spatial dimensions, P2P-voxel uses a flatten operation and a linear layer for subsequent prediction:
\begin{equation}
   \mathcal{M}_{t-1,t} = {\rm MLP}({\rm Linear}({\rm Flatten}(\mathcal{F}_{fusion}^{pc}))).
    \label{eq8}
\end{equation}

\begin{table*}[t]
\caption{Model configurations of our P2P-point and P2P-voxel. The inputs for the two versions are $\mathcal{P}_{t-1/t}^{crop}\in\mathbb{R}^{1024\times3}$ and $\mathcal{V}^{crop}_{t-1/t}\in\mathbb{R}^{128\times 128\times 20}$. ``Sp.'' and ``Ch.'' refer to spatial and channel dimensions, respectively. ``$[k/k\times k,c,s]$'' means convolution layers with kernel size $k/k\times k$, output channel $c$ and stride $s$. Each block ``$[k,c,s]\times n$'' in the neck part of P2P-point is followed by a permutation operation. We report tracking speeds on a single NVIDIA RTX 3090/GTX 1080Ti.}
\centering
\fontsize{7.5pt}{3.5mm}\selectfont
\setlength{\tabcolsep}{0.7mm}{
\resizebox{1.5\columnwidth}{!}{
\begin{tabular}{c|c|c|c|c}
\hline
\multicolumn{1}{l|}{}     
& Output Size
& P2P-point                                                          
& Output Size 
& P2P-voxel \\ 
\hline
Backbone & $\begin{array}{c} \mathcal{F}_{t-1}/\mathcal{F}_{t}: \\1\times1024
\end{array}$ & Point. Embed.&$\begin{array}{c} \mathcal{F}_{t-1}/\mathcal{F}_{t}: \\16\times16\times128\end{array}$ &Voxel. Embed.\\
\hline

\multirow{7}{*}{Neck}    
& $\begin{array}{c} 2\times1024 \end{array}$
& Sp. Concat.                                                     
& $\begin{array}{c} 16\times16\times256 \end{array}$                   
& Ch. Concat.\\ 
\cmidrule{2-5} 

& $\begin{array}{c} 64\times1024 \end{array}$

& \begin{tabular}[c]{@{}c@{}} $\begin{bmatrix}
\!1, 64, 1\! \\
\end{bmatrix} \times 2$ \\$\begin{bmatrix}
\!1, 1024, 1\! \\
\end{bmatrix} \times 2$\end{tabular} 
&$\begin{array}{c} 16\times16\times256 \end{array}$
&      
$\begin{bmatrix}
\!3\times3, 256, 1\! \\
\end{bmatrix} \times 3$
\\ \cmidrule{2-5}

& $\begin{array}{c} 128\times1024 \end{array}$
& \begin{tabular}[c]{@{}c@{}} $\begin{bmatrix}
\!1, 128, 1\! \\
\end{bmatrix} \times 2$\\$\begin{bmatrix}
\!1, 1024, 1\! \\
\end{bmatrix} \times 2$\end{tabular} 
&$\begin{array}{c} 8\times8\times512 \end{array}$
& \begin{tabular}[c]{@{}c@{}} $\begin{bmatrix}
\!3\times3, 512, 2\! \\
\end{bmatrix} \times 1$ \\$\begin{bmatrix}
\!3\times3, 512, 1\! \\
\end{bmatrix} \times 2$\end{tabular} 
\\ \cmidrule{2-5}

& $\begin{array}{c} 256\times1024 \end{array}$
& \begin{tabular}[c]{@{}c@{}} $\begin{bmatrix}
\!1, 256, 1\! \\
\end{bmatrix} \times 2$ \\$\begin{bmatrix}
\!1, 1024, 1\! \\
\end{bmatrix} \times 2$\end{tabular} 
&$\begin{array}{c} 4\times4\times1024 \end{array}$
& \begin{tabular}[c]{@{}c@{}} $\begin{bmatrix}
\!3\times3, 1024, 2\! \\
\end{bmatrix} \times 1$ \\$\begin{bmatrix}
\!3\times3, 1024, 1\! \\
\end{bmatrix} \times 2$\end{tabular} 
\\ \hline

\multirow{5}{*}{Head}    
& $\begin{array}{c} 1\times1024 \end{array}$
&  Maxpooling
& $\begin{array}{c} 1\times1024 \end{array}$                   
& Flatten $\&$ Linear
\\ \cmidrule{2-5} 
                           
& $\begin{array}{c} 1\times4 \end{array}$ 
& $\begin{bmatrix}
\!1, 512, 1\! \\
\!1, 256, 1\! \\
\!1, 128, 1\! \\
\!1, 4, 1\! \\
\end{bmatrix} \times1$
& $\begin{array}{c} 1\times4 \end{array}$ 
& $\begin{bmatrix}
\!1, 512, 1\! \\
\!1, 256, 1\! \\
\!1, 128, 1\! \\
\!1, 4, 1\! \\
\end{bmatrix} \times1$
\\ \hline
\multirow{3}{*}{}
& Params & 7.39 M & Params & 32.00 M\\ \cmidrule{2-5}
& FLOPs & 1.38 G & FLOPs & 1.23 G\\ \cmidrule{2-5}
& Speed & 107/54 Fps & Speed &  71/30 Fps\\ \hline
\end{tabular}
}}
\label{table1}
\end{table*}

\subsection{Effectiveness Analysis}
\label{sec3.6}
\noindent\textbf{Deep Tracking Framework} 
Recent years, the appearance matching~\citep{sc3d,p2b} framework has attracted significant attention. However, its template matching mechanism tend to interfere with target's motion cues, thereby requiring complex structures for precise target localization. For example, the most recent appearance matching based trackers CXTrack~\citep{cxtrack} and MBPTrack \citep{mbptrack} achieve advanced performance contributed to a series of prior structural designs, including a target-centric transformer, a box-prior localization network and so on. The motion-centric framework, represented by M$^2$Track~\citep{m2track}, appears to be a more promising approach. It fully exploits the motion cues of the target between consecutive frames, achieving state-of-the-art performance with an interpretable pipeline. Even better, our proposed P2P framework integrates part-to-part motion modeling, enabling the direct inference of accurate relative motion without relying on architectural components, such as segmentation module, two-stage design found in M$^2$Track. The effectiveness and potential of fusing each corresponding part between spatial structures of consecutive point clouds have been demonstrated in the experiment section. More details are as follows.

\begin{figure*}[t]
  \centering
   \includegraphics[width=1.0\linewidth]{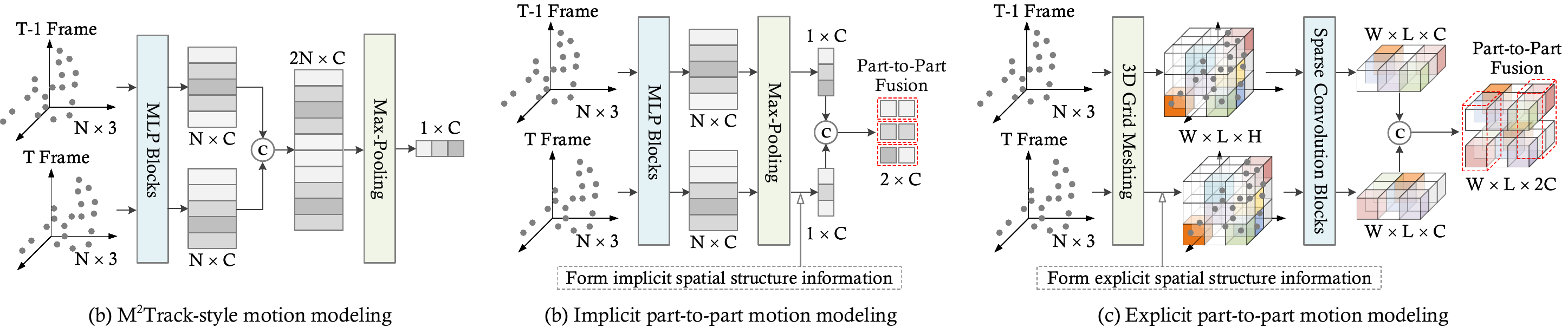}
   \vspace{-10pt}
   \caption{Comparison between M$^2$Track-style motion modeling (a) and part-to-part motion modeling (b)\&(c): (a) M$^2$Track fuse the previous and current frames only by max-pooling operation without learning fine-grained information changes between the two frames. (b) P2P-point implicitly encodes spatial structure information into each feature channel, thus each feature channel can be considered as a implicit part. (c) P2P-voxel enables explicit spatial structure information due to 3D grids, thus each feature spatial is a explicit part.}
   \label{fig6}
\end{figure*}

\noindent\textbf{M$^2$Track v.s. P2P.}
Let $\mathcal{O}_{t-1}$ and $\mathcal{O}_t$ be the information bodies of the two-frame point clouds, which contain a total of $n$ information parts:
\begin{equation}
   \mathcal{O}_{t-1}~{\rm or}~\mathcal{O}_t\in\{o^1,o^2,...,o^n\}.
    \label{eq9}
\end{equation}
As shown in Fig.~\ref{fig6}, M$^2$Track models target motion cues by fusing all information parts:
\begin{equation}
F_{m}(o^1_{t-1},...,o^n_{t-1},o^1_{t},...,o^n_{t}).
    \label{eq10}
\end{equation}
In contrast, our proposed P2P framework perform motion modeling in a part-to-part manner:
\begin{equation}
   F_{m}^{pp}(o^1_{t-1,t},...,o^n_{t-1,t}).
    \label{eq11}
\end{equation}
For P2P-point, $n$ is $C$, and each part feature $\in1\times1$, where part-to-part refers to channel-to-channel. For P2P-voxel, $n$ is $W\cdot L$, and each part feature $\in1\times C$, where part-to-part refers to spatial-to-spatial. As a result, more fine-grained motion cues in each part are modeled, guiding a more accurate and robust motion prediction. Benefiting from this, our methods work in a one-stage manner without extra segmentation and two-stage refinement modules used in M$^2$Track~\citep{m2track}, while demonstrating state-of-the-art performance and high speed.

\noindent\textbf{P2P-point v.s. P2P-voxel.} 
The superior performance of the proposed framework stems from its part-to-part motion modeling, which facilitates robust and accurate motion cues for tracking, as demonstrated in our manuscript. A critical component of this process is part-to-part fusion, which directly impacts the quality of motion cues and thus affects the overall performance. Compared to P2P-point, P2P-voxel has inherent advantages in fusing corresponding parts between consecutive two-frame point clouds. Through voxel-based representation, P2P-voxel preserves the spatial structure of point clouds, enabling explicit part-to-part fusion, where each pixel explicitly represents a distinct part. This precise correspondence allows for more accurate motion modeling. In contrast, P2P-point utilizes a point-based representation, relying on pooling operations to maintain the permutation invariance of point clouds. This approach embeds the spatial structure into a 1D semantic feature, where each channel implicitly represents a part of the overall point cloud information. As a result, P2P-point achieves implicit part-to-part fusion between consecutive point clouds. While both point and voxel representation-based models have their merits, explicit part-to-part fusion used in P2P-voxel allows for a more precise alignment and interaction between corresponding parts of consecutive point clouds, thereby outperforming the implicit approach of P2P-point. Therefore, P2P-voxel presents performance advantage over P2P-point.

\begin{table*}[t]
\centering
\caption{Backbone architectures of our customized PointNet and VoxelNet. ``$[k,c,s]\times n$" means $n$ 1D convolution layers with kernel size $k$, output channel $c$ and stride $s$, while ``$[k\times k\times k,c,s]\times n$" means $n$ 3D sparse convolution layers with kernel size $k\times k\times k$, output channel $c$ and stride $s$. ``\#$n$'' represents the $n$-th stage within the backbones.}
\fontsize{7.5pt}{3.5mm}\selectfont
\setlength{\tabcolsep}{0.7mm}{
\resizebox{1.5\columnwidth}{!}{
\begin{tabular}{c|c|c|c|c}
\hline
\multicolumn{1}{l|}{} 
& Output Size
& Point. Embed                                                          
& Output Size 
& Voxel. Embed \\
\hline
\#1 
& $\begin{array}{c} 1024\times64 \end{array}$
& $\begin{bmatrix} 
\!1, 64, 1\! \\
\end{bmatrix} \times 2$
&$\begin{array}{c} 64\times64\times10\times32 \end{array}$
& \begin{tabular}[c]{@{}c@{}} $\begin{bmatrix}
\!3\times3\times3, 16, 1\! \\
\end{bmatrix} \times 2$ \\$\begin{bmatrix}
\!3\times3\times3, 32, 1\! \\
\end{bmatrix} \times 1$\end{tabular} \\
\hline
\#2
& $\begin{array}{c} 1024\times128 \end{array}$
& $\begin{bmatrix}
\!1, 128, 1\! \\
\end{bmatrix} \times 1$
&$\begin{array}{c} 32\times32\times5\times64 \end{array}$
& \begin{tabular}[c]{@{}c@{}} $\begin{bmatrix}
\!3\times3\times3, 32, 1\! \\
\end{bmatrix} \times 2$ \\$\begin{bmatrix}
\!3\times3\times3, 64, 1\! \\
\end{bmatrix} \times 1$\end{tabular} \\
\hline
\#3 
& $\begin{array}{c} 1024\times1024 \end{array}$
& $\begin{bmatrix}
\!1, 1024, 1\! \\
\end{bmatrix} \times 1$
&$\begin{array}{c} 16\times16\times3\times128 \end{array}$
& \begin{tabular}[c]{@{}c@{}} $\begin{bmatrix}
\!3\times3\times3, 64, 1\! \\
\end{bmatrix} \times 2$ \\$\begin{bmatrix}
\!3\times3\times3, 128, 1\! \\
\end{bmatrix} \times 1$\end{tabular} \\
\hline
 
& $\begin{array}{c} 1\times1024 \end{array}$
& Maxpooling
&$\begin{array}{c} 16\times16\times128 \end{array}$
& Flatten \& Linear \\
\hline
\end{tabular}
}}
\label{table_backbone}
\end{table*}

\section{Experiment}

\subsection{Implementation Details}

\noindent\textbf{Model Hyper-parameters.} 
We define search regions for two consecutive frames at timestamps $t-1$ and $t$ by centering on the previous target bounding box within a 3D space. The ranges are defined as [(-4.8,4.8),(-4.8,4.8),(-1.5,1.5)] and [(-1.92,1.92),(-1.92,1.92),(-1.5,1.5)] to contain relevant points for cars and humans, respectively. To be a fair comparison, we follow existing methods~\citep{p2b,m2track} to sub-sample 1024 points using farthest point sampling~\citep{pointnet++} within the derived search regions as inputs for P2P-point. For P2P-voxel, we do not down-sample point clouds. Instead, we divide the search regions into 128$\times$128$\times$20 voxel grids, which serve as inputs for the model. More model details can be found in Tab.~\ref{table1}. 

Tab.~\ref{table_backbone} presents the backbone architectures utilized for P2P-point and P2P-voxel, \textit{i.e.}, ``Point. Embed'' and ``Voxel. Embed'' in Tab.~\ref{table1}of the main manuscript. We modify the PointNet~\citep{pointnet} and VoxelNet~\citep{voxelnet} as our backbones to extract features. Specifically, we remove task layers and T-Net of PointNet, using only 1D convolution layers with batch normalization (BN) and rectified linear unit (ReLU) function to extract point features. Similarly, we remove the task layers of VoxelNet and adopt 3D sparse convolution blocks without residual structure to extract the features of input voxels.

\noindent\textbf{Data Augmentation.} 
To enhance the model's robustness, we introduce simulated test errors for both frame $t-1$ and $t$ during the training. These simulations involve the random horizontal flipping of points and bounding boxes of the targets, along with uniform rotations around their vertical axis within [-5$^\circ$,5$^\circ$]. To boost the model's accuracy, we incorporate various simulated motion patterns in the current frame $t$ by adding random translations to the targets along the $xyz$ axes. The translations are modeled using a Gaussian distribution with parameters [$\mu$, $\sigma$]. For $x$, $y$, and $z$ axes, the mean $\mu$ and variance $\sigma$ values are configured as [0,0.3], [0,0.1], and [0,0.1], respectively.

\begin{algorithm*}[t]
    \caption{Tracking process of P2P-point}
    \label{al1}
    \KwIn{A $T$-frame point cloud sequence ${\{\mathcal{P}_t\}_{t=1}^{T}}$; A target bounding box $\mathcal{B}_1=(x_1,y_1,z_1,w_1,h_1,l_1,\theta_1)$ in the $1$-st frame of the given sequence.} 
    \KwOut{A series of predicted bounding boxes $\{\mathcal{B}_t=(x_t,y_t,z_t,w_t,h_t,l_t,\theta_t)\}_{t=2}^{T}$ from $2$-nd frame to $T$-th frame.}  
    \For{$t=2$ to $T$}  
    { 
      Crop previous search region $\mathcal{P}_{t-1}^{crop}\in\mathbb{R}^{1024\times3}$ from point cloud $\mathcal{P}_{t-1}$ in the ($t-1$)-th frame;\\
      Crop current search region $\mathcal{P}_{t}^{crop}\in\mathbb{R}^{1024\times3}$ from point cloud $\mathcal{P}_{t}$ in the $t$-th frame;\\
      Use a weight-shared PointNet to encode $\mathcal{P}_{t-1}^{crop}$ and $\mathcal{P}_{t}^{crop}$ to embeddings $\mathcal{F}_{t-1}\in\mathbb{R}^{1\times1024}$ and $\mathcal{F}_{t}\in\mathbb{R}^{1\times1024}$;\\
      Fuse $\mathcal{F}_{t-1}$ and $\mathcal{F}_{t}$ in a spatial-aligned manner to obtain fused feature $\mathcal{F}_{fusion}^{sa}\in\mathbb{R}^{256\times1024}$;\\
      Obtain 4-DOF relative motion $\mathcal{M}_{t-1,t}=(\Delta x_t,\Delta y_t,\Delta z_t,\Delta \theta_t)$ by applying a MLP based prediction head;\\
      Transform previous $\mathcal{B}_{t-1}$ using $\mathcal{M}_{t-1,t}$ to output $\mathcal{B}_t$ in current frame.\\
    } 
    \textbf{Return} $\{\mathcal{B}_t\}_{t=2}^{T}$.
\end{algorithm*}

\begin{algorithm*}[!t]
    \caption{Tracking process of P2P-voxel}
    \label{al2}
    \KwIn{A $T$-frame point cloud sequence ${\{\mathcal{P}_t\}_{t=1}^{T}}$; A target bounding box $\mathcal{B}_1=(x_1,y_1,z_1,w_1,h_1,l_1,\theta_1)$ in the $1$-st frame of the given sequence.} 
    \KwOut{A series of predicted bounding boxes $\{\mathcal{B}_t=(x_t,y_t,z_t,w_t,h_t,l_t,\theta_t)\}_{t=2}^{T}$ from $2$-nd frame to $T$-th frame.}  
    \For{$t=2$ to $T$}  
    {  
      Crop previous search region $\mathcal{V}_{t-1}^{crop}\in\mathbb{R}^{128\times128\times20}$ from point cloud $\mathcal{P}_{t-1}$ in the ($t-1$)-th frame;\\
      Crop current search region $\mathcal{V}_{t}^{crop}\in\mathbb{R}^{128\times128\times20}$ from point cloud $\mathcal{P}_{t}$ in the $t$-th frame;\\
      Use a weight-shared VoxelNet to encode $\mathcal{V}_{t-1}^{crop}$ and $\mathcal{V}_{t}^{crop}$ to $\mathcal{F}_{t-1}\in\mathbb{R}^{16\times16\times128}$ and $\mathcal{F}_{t}\in\mathbb{R}^{16\times16\times128}$;\\
      Fuse $\mathcal{F}_{t-1}$ and $\mathcal{F}_{t}$ in a spatial-aligned manner to obtain fused feature $\mathcal{F}_{fusion}^{sa}\in\mathbb{R}^{4\times4\times1024}$;\\
      Obtain 4-DOF relative motion $\mathcal{M}_{t-1,t}=(\Delta x_t,\Delta y_t,\Delta z_t,\Delta \theta_t)$ by applying a 2D Conv based prediction head;\\
      Transform previous $\mathcal{B}_{t-1}$ using $\mathcal{M}_{t-1,t}$ to output $\mathcal{B}_t$ in current frame.\\
    } 
    \textbf{Return} $\{\mathcal{B}_t\}_{t=2}^{T}$.
\end{algorithm*}

\noindent\textbf{Training \& Inference Details.} We train our tracking models using the AdamW optimizer on a Tesla A100 GPU, with a batch size of 128. The initial learning rate is set to 1$e$-4 and is decayed by a factor of 5 every 20 epochs. P2P-point and P2P-voxel take $\sim$ 50 and 90 seconds, respectively, per training epoch on the KITTI Car category. During the inference stage, the models predict a target frame-by-frame in a continuous sequence of point clouds, given the target bounding box in the initial frame. Alg.~\ref{al1} and \ref{al2} present the entire tracking process on a point cloud scene of P2P-point and P2P-voxel, respectively. 

\begin{table*}[t]
\caption{Comparisons with state-of-the-art methods on KITTI dataset~\citep{kitti}. The best three results are colored in \textcolor{red}{red}, \textcolor{blue}{blue} and \textcolor{green}{green}. Success/Precision are used for evaluation. We use ``$\ast$'' to represent base tracking frameworks in the 3D SOT community.}
\centering
    \resizebox{1\textwidth}{!}{
    \normalsize
    \begin{tabular}{c|c|c|ccccc|cc}
          \toprule[0.4mm]
          \rowcolor{black!10} & & & Car & Pedestrian &  Van & Cyclist& Mean && \\
          \rowcolor{black!10} \multirow{-2}{*}{Paradigm}&\multirow{-2}{*}{Tracker} & \multirow{-2}{*}{Source} & [6,424]&[6,088] & [1,248] & [308] & [14,068] & \multirow{-2}{*}{Hardware} & \multirow{-2}{*}{Fps}\\
          
          \midrule
          \multirow{4}{*}{Motion}
          &\textbf{P2P-voxel}  & Ours& \textcolor{red}{73.6}/\textcolor{red}{85.7}&\textcolor{red}{69.6}/\textcolor{red}{94.0}&\textcolor{red}{70.3}/\textcolor{red}{83.9}&\textcolor{green}{75.5}/\textcolor{blue}{94.6}&\textcolor{red}{71.7}/\textcolor{red}{89.4} & RTX 3090&71\\
         &\textbf{P2P-point} & Ours&68.8/81.7&62.7/89.1&\textcolor{blue}{65.4}/\textcolor{blue}{80.1}&74.8/\textcolor{red}{94.8}&	66.2/\textcolor{green}{85.4} & RTX 3090&107\\
          & M$^2$Track++~\citep{zheng2023effective} & TPAMI'23&71.1/82.7  & 61.8/88.7 & \textcolor{green}{62.8}/\textcolor{green}{78.5} & \textcolor{blue}{75.9}/94.0 & 66.5/85.2 & Tesla V100&57\\
          & M$^2$Track$^\ast$~\citep{m2track} & CVPR'22&65.5/80.8   & 61.5/88.2 & 53.8/70.7 & 73.2/93.5 & 62.9/83.4 & Tesla V100&57\\
          \midrule
          & PTTR++~\citep{luo2024exploring}& TPAMI'24&73.4/84.5 &55.2/84.7 &55.1/62.2& 71.6/92.8& 63.9/82.8& Tesla V100 & 43\\
          &MBPTrack~\citep{mbptrack} & ICCV'23 & \textcolor{green}{73.4}/\textcolor{green}{84.8} &\textcolor{blue}{68.6}/\textcolor{blue}{93.9} &61.3/72.7& \textcolor{red}{76.7}/\textcolor{green}{94.3}& \textcolor{blue}{70.3}/\textcolor{blue}{87.9} & RTX 3090 & 50 \\
           &SyncTrack~\citep{synctrack} &ICCV'23 &73.3/\textcolor{blue}{85.0} & 54.7/80.5&60.3/70.0& 73.1/93.8 & 64.1/81.9& TITAN RTX& 45\\
          & CXTrack~\citep{cxtrack} & CVPR'23& 69.1/81.6 &\textcolor{green}{67.0}/\textcolor{green}{91.5}& 60.0/71.8& 74.2/94.3& \textcolor{green}{67.5}/85.3& RTX 3090 & 29\\
           & CorpNet~\citep{corpnet} & CVPRw'23&\textcolor{blue}{73.6}/84.1 &55.6/82.4&58.7/66.5 &74.3/94.2 & 64.5/82.0& TITAN RTX& 36\\
           & OSP2B~\citep{osp2b}& IJCAI'23&67.5/82.3   & 53.6/85.1 & 56.3/66.2 & 65.6/90.5 & 60.5/82.3 & GTX 1080Ti & 34\\
           Appearance& GLT-T~\citep{glt}&AAAI'23 &68.2/82.1   & 52.4/78.8 & 52.6/62.9 & 68.9/92.1 & 60.1/79.3 & GTX 1080Ti &30\\
          Matching &CMT~\citep{cmt} & ECCV'22 &70.5/81.9 &49.1/75.5& 54.1/64.1& 55.1/82.4 &59.4/77.6 & GTX 1080Ti & 32\\
           &STNet~\citep{stnet} & ECCV'22 &72.1/84.0 &49.9/77.2& 58.0/70.6& 73.5/93.7& 61.3/80.1& TITAN RTX& 35\\
           & PTTR~\citep{pttr}& CVPR'22&65.2/77.4   & 50.9/81.6 & 52.5/61.8 & 65.1/90.5 & 57.9/78.2 & Tesla V100 & 50\\
           & V2B~\citep{v2b}&NeurIPS'21 &70.5/81.3   & 48.3/73.5 & 50.1/58.0 & 40.8/49.7 & 58.4/75.2 &  TITAN RTX& 37\\
          & BAT~\citep{bat}&ICCV'21& 60.5/77.7 &42.1/70.1 & 52.4/67.0 &33.7/45.4& 51.2/72.8 & RTX 2080 & 57 \\
          &  MLVSNet~\citep{mlvsnet}&ICCV'21 &56.0/74.0   & 34.1/61.1 & 52.0/61.4 & 34.4/44.5 & 45.7/66.6 & GTX 1080Ti & 70\\
           &P2B$^\ast$~\citep{p2b}& CVPR'20& 56.2/72.8 & 28.7/49.6 & 40.8/48.4 & 32.1/44.7 & 42.4/60.0 & GTX 1080Ti & 40\\
          & SC3D$^\ast$~\citep{sc3d}& CVPR'19 &41.3/57.9   & 18.2/37.8 & 40.4/47.0 & 41.5/70.4 & 31.2/48.5 & GTX 1080Ti & 2\\
          \bottomrule[0.4mm]
    \end{tabular}}
\label{table2}
\end{table*}

\subsection{Experiment Setting}

\noindent\textbf{Datasets.} 
We conduct comprehensive experiments on three widely used datasets, including KITTI~\citep{kitti}, NuScenes~\citep{Nuscenes} and Waymo Open Dataset (WOD)~\citep{waymo}. KITTI consists of 21 training and 29 test point cloud sequences, respectively. Due to the unavailability of test labels, we divide the training sequences into training set [0-17), validation set [17-19) and test set [19-21). Compared to KITTI, NuScenes and WOD provide more challenging and large-scale scenes. NuScenes contains 700 and 150 scenes for training and testing, while WOD comprises 1121 tracklets categorized into easy, medium and hard sub-sets based on the point cloud sparsity.

\noindent\textbf{Evaluation Metrics.} 
Following common practice, we employ One Pass Evaluation (OPE)~\citep{otb2013} to evaluate tracking performance using both Success and Precision metrics. Success calculates the intersection over union (IOU) between the predicted bounding box and the ground truth one, while Precision assesses the distance between the centers of the two corresponding bounding boxes.

\begin{table*}[t]
\caption{Comparisons with state-of-the-art methods on Waymo Open Dataset~\citep{waymo}.}
\centering
    \resizebox{1.0\textwidth}{!}{
    \normalsize
    \begin{tabular}{c|cccc|cccc|c}
          \toprule[0.4mm]
         \rowcolor{black!10} & \multicolumn{4}{c|}{Vehicle} & \multicolumn{4}{c|}{Pedestrian } & \\
        \rowcolor{black!10}  & Easy& Medium & Hard & Mean & Easy & Medium & Hard & Mean & \\
       \rowcolor{black!10} \multirow{-3}{*}{Tracker}  & [67,832]&[61,252]&[56,647]&[185,731]& [85,280]&[82,253]&[74,219]&[241,752] & \multirow{-3}{*}{Mean}\\
          \midrule
          \textbf{P2P-voxel}  &  \textcolor{blue}{66.2}/\textcolor{blue}{73.8}&\textcolor{blue}{57.8}/\textcolor{blue}{67.0}&\textcolor{blue}{56.8}/\textcolor{blue}{68.1} &\textcolor{blue}{60.0}/\textcolor{blue}{69.1} &\textcolor{red}{43.7}/\textcolor{red}{65.2} &\textcolor{red}{36.4}/\textcolor{red}{57.1}&\textcolor{red}{31.3}/\textcolor{red}{51.0} &\textcolor{red}{37.4}/\textcolor{red}{58.1} &\textcolor{red}{47.2}/\textcolor{red}{62.9}\\
         \textbf{P2P-point} &  61.3/68.2&53.4/61.8& 51.9/61.5 &55.3/63.4 &\textcolor{green}{37.0}/\textcolor{green}{55.9}&\textcolor{green}{31.1}/\textcolor{green}{48.5}&\textcolor{green}{28.5}/\textcolor{green}{45.3}&\textcolor{green}{32.4}/\textcolor{green}{50.1} &\textcolor{green}{42.3}/55.9\\
          \midrule
          MBPTrack~\citep{mbptrack} & \textcolor{red}{68.5}/\textcolor{red}{77.1} &\textcolor{red}{58.4}/\textcolor{red}{68.1}& \textcolor{red}{57.6}/\textcolor{red}{69.7}& \textcolor{red}{61.9}/\textcolor{red}{71.9} &\textcolor{blue}{37.5}/\textcolor{blue}{57.0} &\textcolor{blue}{33.0}/\textcolor{blue}{51.9}& \textcolor{blue}{30.0}/\textcolor{blue}{48.8}& \textcolor{blue}{33.7}/\textcolor{blue}{52.7} &\textcolor{blue}{46.0}/\textcolor{blue}{61.0}\\
          CXTrack~\citep{cxtrack} & 63.9/71.1 &54.2/62.7& 52.1/63.7 &57.1/66.1& 35.4/55.3 &29.7/47.9& 26.3/44.4 &30.7/49.4& 42.2/\textcolor{green}{56.7} \\
          STNet~\citep{stnet}& \textcolor{green}{65.9}/\textcolor{green}{72.7}& \textcolor{green}{57.5}/\textcolor{green}{66.0}& \textcolor{green}{54.6}/\textcolor{green}{64.7} &\textcolor{green}{59.7}/\textcolor{green}{68.0}& 29.2/45.3& 24.7/38.2& 22.2/35.8 &25.5/39.9 &40.4/52.1 \\
         V2B~\citep{v2b}& 64.5/71.5 & 55.1/63.2 & 52.0/62.0 & 57.6/65.9 & 27.9/43.9 & 22.5/36.2 & 20.1/33.1 & 23.7/37.9 & 38.4/50.1\\
          BAT~\citep{bat}& 61.0/68.3 & 53.3/60.9 & 48.9/57.8 & 54.7/62.7 & 19.3/32.6 & 17.8/29.8 & 17.2/28.3 & 18.2/30.3 & 34.1/44.4\\
          P2B~\citep{p2b}& 57.1/65.4 & 52.0/60.7 & 47.9/58.5 & 52.6/61.7 & 18.1/30.8 & 17.8/30.0 & 17.7/29.3 & 17.9/30.1 & 33.0/43.8\\ 
          \bottomrule[0.4mm]
    \end{tabular}}
\label{table3}
\end{table*}

\subsection{Comparison with State-of-the-art Trackers}

\noindent\textbf{Results on KITTI.} 
We present a comprehensive comparison between the proposed methods and the previous state-of-the-art methods, including recent trackers like SyncTrack~\citep{synctrack} and MBPTrack~\citep{mbptrack} on the KITTI dataset. As shown in Tab.~\ref{table2}, P2P-voxel exhibits superior performance across various categories, achieving the highest mean Success and Precision rates of 71.7\% and 89.4\%, respectively. Moreover, P2P-voxel outperforms the previous leading method, \textit{i.e.}, MBPTrack by 1.4\%, while demonstrating notable advantages in terms of running speed. Compared to the motion track M$^2$Track~\citep{m2track}, P2P-point achieves remarkable performance improvements in all categories. This proves the effectiveness of our strong framework, which performs part-to-part fusion between the spatial structures of consecutive point clouds for motion modeling. By leveraging the voxel-based representation to enable explicit part-to-part fusion of point cloud spatial structures, P2P-voxel further improves the tracking performance by a significant margin.

\noindent\textbf{Results on WOD.} 
To validate the generalization ability of the proposed methods, we conduct an evaluation by applying the Car and Pedestrian models trained on KITTI dataset to the WOD dataset, following comment setting~\citep{v2b,mbptrack}. As presented in Tab.~\ref{table3}, our P2P-voxel outperforms other comparison methods, especially in the Pedestrian category, indicating strong generalization of our proposed framework to unseen scenes. Moreover, although P2P-point incorporates implicit part-to-part motion modeling through point-based representation that poses more stringent generalization conditions, it still achieves competitive performance. 

\begin{table*}[t]
\caption{Comparisons with state-of-the-art methods on NuScenes dataset~\citep{Nuscenes}.}
\centering
    \resizebox{1.0\textwidth}{!}{
    \normalsize
    \begin{tabular}{c|ccccc|c|c}
          \toprule[0.4mm]
          \rowcolor{black!10}  & Car & Pedestrian &  Truck & Trailer & Bus & Mean & Mean by\\
          \rowcolor{black!10} \multirow{-2}{*}{Track}  & [64,159]&[33,227] & [13,587] & [3,352] & [2,953] & [117,278] &Category\\
          \midrule
           \textbf{P2P-voxel} & \textcolor{red}{65.15}/\textcolor{red}{72.90}&\textcolor{red}{46.43}/\textcolor{red}{75.08}&\textcolor{red}{64.96}/\textcolor{red}{65.96}&\textcolor{red}{70.46}/\textcolor{red}{66.86}&\textcolor{red}{59.02}/\textcolor{red}{56.56}	&\textcolor{red}{59.84}/\textcolor{red}{72.13}&\textcolor{red}{61.04}/\textcolor{red}{67.44}\\
          \textbf{P2P-point}  &  \textcolor{green}{62.14}/\textcolor{green}{68.45}& \textcolor{green}{39.68}/\textcolor{green}{65.59}&\textcolor{blue}{62.50}/\textcolor{blue}{63.44}&\textcolor{blue}{69.04}/\textcolor{blue}{65.14}&\textcolor{blue}{57.90}/\textcolor{blue}{55.46}& \textcolor{green}{55.92}/\textcolor{green}{66.64}&\textcolor{blue}{58.25}/\textcolor{green}{63.62}\\
           M$^2$Track~\citep{m2track}&55.85/65.09& 32.10/60.92 &57.36/59.54& 57.61/58.26& 51.39/51.44& 49.23/62.73& 50.86/59.05\\
          \midrule
          MBPTrack~\citep{mbptrack}  &\textcolor{blue}{62.47}/\textcolor{blue}{70.41} &\textcolor{blue}{45.32}/\textcolor{blue}{74.03} & \textcolor{green}{62.18}/\textcolor{green}{63.31} & \textcolor{green}{65.14}/\textcolor{green}{61.33}&  \textcolor{green}{55.41}/\textcolor{green}{51.76} &\textcolor{blue}{57.48}/\textcolor{blue}{69.88}&  \textcolor{green}{58.10}/\textcolor{blue}{64.19}\\
          GLT-T~\citep{glt}&48.52/54.29&31.74/56.49& 52.74/51.43&57.60/52.01& 44.55/40.69&44.42/54.33 & 47.03/50.98\\
           PTTR~\citep{pttr}& 51.89/58.61& 29.90/45.09& 45.30/44.74 &45.87/38.36& 43.14/37.74& 44.50/52.07& 43.22/44.91\\
          BAT~\citep{bat}& 40.73/43.29 &28.83/53.32& 45.34/42.58& 52.59/44.89 &35.44/28.01 &38.10/45.71&40.59/42.42\\
          PTT~\citep{ptt} &41.22/45.26& 19.33/32.03&50.23/48.56&51.70/46.50&39.40/36.70 & 36.33/41.72 & 40.38/41.81\\
           P2B~\citep{p2b}& 38.81/43.18 &28.39/52.24& 42.95/41.59& 48.96/40.05& 32.95/27.41& 36.48/45.08& 38.41/40.90\\
          SC3D~\citep{sc3d} &22.31/21.93& 11.29/12.65& 30.67/27.73& 35.28/28.12& 29.35/24.08& 20.70/20.20&25.78/22.90\\
          \bottomrule[0.4mm]
    \end{tabular}}
\label{table4}
\end{table*}

\noindent\textbf{Results on NuScenes.} We further conduct comparative experiments on the more challenging NuScenes dataset. Here, we select existing methods that have reported relevant results as comparisons, including SC3D~\citep{sc3d}, P2B~\citep{p2b}, PTT~\citep{ptt}, BAT~\citep{bat}, PTTR~\citep{ptt}, GLT-T~\citep{glt}, M$^2$Track~\citep{m2track} and MBPTrack~\citep{mbptrack}. Our P2P-voxel performs better than other methods across all categories, as shown in Tab.~\ref{table4}. Notably, P2P-voxel and P2P-point exhibit considerable leading performance in the Truck, Trailer and Bus categories. These results imply that our proposed P2P framework has the capacity to deliver exceptional model performance without a large-sample training. In addition, given the complex scenes in the NuScenes dataset, the outstanding performance demonstrates the potential of our framework for practical applications.

\begin{figure*}[t]
    \centerline{\includegraphics[width=1.05\linewidth]{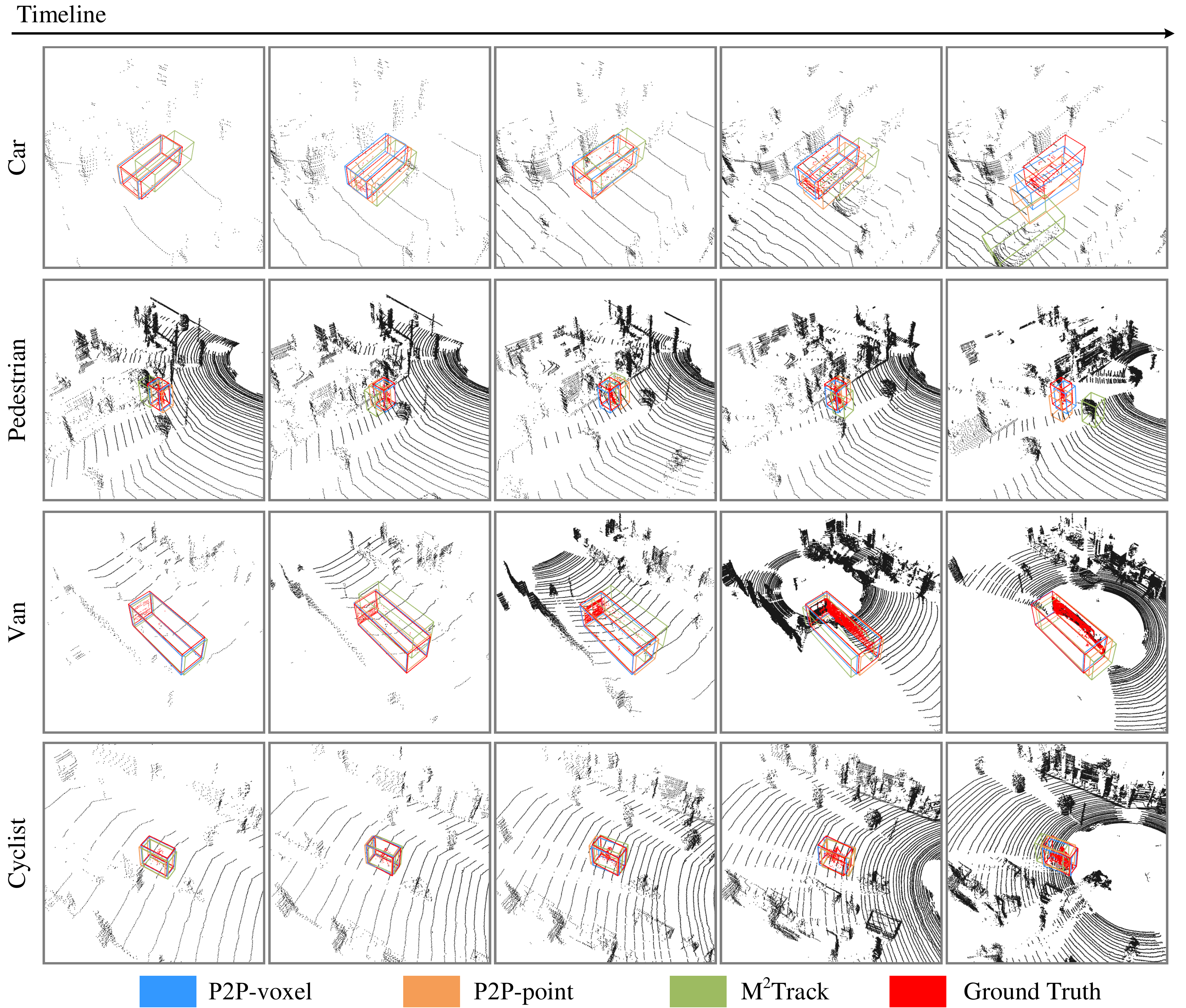}}
    \caption{Visual results of our P2P-voxel/-point and M$^2$Track on the point cloud sequences of Car, Pedestrian, Van, and Cyclist categories (from top to bottom). The \textbf{\textcolor{red}{red}} points and box are the foreground points and ground truth of targets. The \textbf{\textcolor{lightblue}{blue}}, \textbf{\textcolor{lightyellow}{yellow}} and \textbf{\textcolor{lightgreen}{green}} boxes denote the prediction results by our P2P-voxel, P2P-point and M$^2$Track, respectively.}
    \label{fig13}
\end{figure*}

\noindent\textbf{Qualitative Comparison.}
To visually demonstrate the effectiveness of the proposed models P2P-voxel and P2P-point, some visualization comparisons are conducted against the existing motion-centric method M$^2$Track~\citep{m2track} on the Car, Pedestrian, Van and Cyclist categories from the KITTI~\citep{kitti} dataset. As illustrated in Fig.~\ref{fig13}, our methods exhibit more accurate object tracking in scenes with distractors or sparse contexts.

\subsection{Ablation Studies}

\begin{figure*}[!t]
  \centering
   \includegraphics[width=1.0\linewidth]{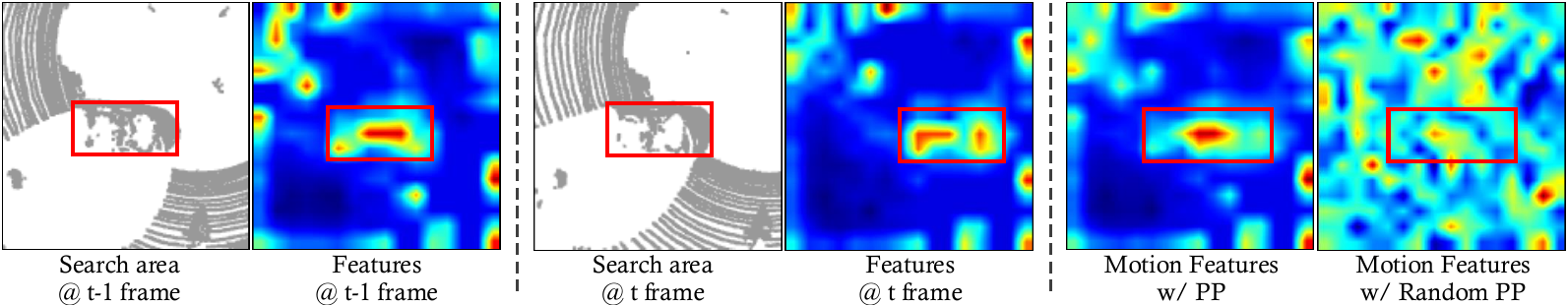}
   \vspace{-10pt}
   \caption{Feature visualization of the motion features after part-to-part motion modeling and random part-to-part motion modeling.}
   \label{fig7}
\end{figure*}

\textbf{Effectiveness of Part-to-Part Motion Modeling.} To analyze our designed part-to-part motion modeling, we first conduct an ablation study on P2P-point (see Fig.~\ref{fig5}) on the KITTI dataset. As reported in Tab.~\ref{table5}, without either part-to-part fusion [\textcircled{3}] or motion modeling in spatial dimension [\textcircled{2}], information changes in the corresponding spatial parts will not be explored, leading to degradation of motion cues, which leads to 10.9\% and 6.8\% decrease in mean Success rate. Moreover, we further remove part-to-part fusion in P2P-voxel by randomly fusing non-corresponding parts [\textcircled{2}]. In such case, the performance degrades remarkably across all categories. Fig.~\ref{fig7} also shows that the features after motion modeling will become cluttered.

\begin{table*}[t]
\caption{Ablation of part-to-part modeling module in P2P-point on KITTI dataset. ``PP'' and ``MM'' denote part-to-part fusion and motion modeling, respectively.}
\centering
    \resizebox{0.825\linewidth}{!}{
    \normalsize
    \begin{tabular}{c|c|cc|ccccc}
          \toprule[0.4mm]
          \rowcolor{black!10} Tracker &\#& PP & Spatial MM & Car & Pedestrian & Van & Cyclist & Mean \\
          \midrule
         \multirow{3}{*}{P2P-point}&\textcircled{1} &\checkmark & \checkmark& 68.8/81.7 &62.7/89.1& 65.4/80.1& 74.8/94.8 & 66.2/85.4 \\
         &\textcircled{2} &\checkmark& & 65.0/77.5&42.6/72.5&61.9/76.6&73.3/94.1 &55.3/75.7\\
          &\textcircled{3}&& \checkmark& 61.7/74.2&55.1/83.2&64.3/76.2&74.3/94.4&59.4/78.8\\
          \midrule
          \multirow{2}{*}{P2P-voxel}&\textcircled{1} &\checkmark & -& 73.6/85.7 &69.6/94.0 &70.3/83.9 &75.5/94.6&71.7/89.4\\
         &\textcircled{2} && -& 57.8/69.5&39.4/65.2&44.3/60.8&61.4/87.8 &48.76/67.3\\
          \bottomrule[0.4mm]
    \end{tabular}}
\label{table5}
\end{table*}

\begin{table*}[!t]
\caption{Ablation of weight-shared feature extractors in P2P-point and P2P-voxel, respectively on KITTI dataset.}
\centering
    \resizebox{0.875\linewidth}{!}{
    \normalsize
    \begin{tabular}{c|c|c|ccccc}
          \toprule[0.4mm]
          \rowcolor{black!5}Tracker&\# &Weight-shared Backbone & Car & Pedestrian & Van & Cyclist & Mean \\
          \midrule
          \multirow{2}{*}{P2P-point}&\textcircled{1} &\checkmark &68.8/81.7 &62.7/89.1& 65.4/80.1& 74.8/94.8 & 66.2/85.4\\
          &\textcircled{2} & &66.1/78.4&53.1/82.7&	64.4/78.3&74.6/94.4&60.6/80.7\\
           \midrule
           \multirow{2}{*}{P2P-voxel}&\textcircled{1} & \checkmark&73.6/85.7 &69.6/94.0 &70.3/83.9 &75.5/94.6&71.7/89.4\\
          &\textcircled{2}& &73.0/84.7&64.4/90.0	&68.6/83.4&74.6/94.6&69.0/87.2\\
          \bottomrule[0.4mm]
    \end{tabular}}
\label{table6}
\end{table*}

\begin{table*}[!t]
\caption{Ablation of variant designs in P2P-voxel on KITTI dataset. ``F$\to$P'' and ``P$\to$F'' denotes ``feature extraction$\to$part-to-part fusion'' and ``part-to-part fusion$\to$feature extraction'', respectively.}
\centering
    \resizebox{0.8\linewidth}{!}{
    \normalsize
    \begin{tabular}{c|c|cc|ccccc}
          \toprule[0.4mm]
          \rowcolor{black!5}Tracker&\# &F$\to$P & P$\to$F& Car & Pedestrian & Van & Cyclist & Mean \\
          \midrule
           \multirow{2}{*}{P2P-voxel}&\textcircled{1} & \checkmark&&73.6/85.7 &69.6/94.0 &70.3/83.9 &75.5/94.6&71.7/89.4\\
          &\textcircled{2}&  & \checkmark &72.7/84.0& 68.5/91.8 &70.2/84.1 &75.3/93.6 &70.9/87.7\\
          \bottomrule[0.4mm]
    \end{tabular}}
\label{table7}
\end{table*}

We then ablate weight-shared feature extractors used in P2P-point and P2P-voxel to investigate part-to-part motion modeling. As shown in Tab.~\ref{table6}, when using non-sharing weighted feature extractor in P2P-point, mean performance degrades significantly. This is because spatial structure information of disordered point clouds is implicitly formed by feature extractor, and unshared parameters extract the features of implicit spatial structure from the consecutive point clouds in a separate semantic space, interfering with the part-to-part fusion. For P2P-voxel, however, explicit spatial structure information is formed before feature extraction, as shown in Fig.~\ref{fig6}. As a result, a relatively slight performance drop is observed, with a Success rate of 0.6\% in Car category. 

In addition, we compare two variant designs of P2P-voxel: ``feature extraction $\to$part-to-part fusion'' (default setting [\textcircled{1}]) v.s. ``part-to-part fusion$\to$feature extraction'' ([\textcircled{2}]) in Tab.~\ref{table7}. Both designs fuse each corresponding part between the explicit spatial structures of consecutive point clouds. Thus, there is a small performance gap, which implies the effectiveness and potential of motion modeling in a part-to-part manner.

\begin{table*}[t]
\caption{Ablation of data augmentation in P2P-point and P2P-voxel, respectively on KITTI~\citep{kitti} dataset.}
\centering
    \resizebox{0.85\linewidth}{!}{
    \normalsize
    \begin{tabular}{c|c|c|ccccc}
    \toprule[0.4mm]
          \rowcolor{black!5}Tracker&\# &Data Augmentation & Car & Pedestrian & Van & Cyclist & Mean \\
          \midrule
          \multirow{2}{*}{P2P-point}&\textcircled{1} & \checkmark&68.8/81.7 &62.7/89.1& 65.4/80.1& 74.8/94.8 & 66.2/85.4\\
          &\textcircled{2} & &66.8/79.0&60.5/86.0&62.9/78.0&74.2/94.6&64.0/82.4\\
           \midrule
           \multirow{2}{*}{P2P-voxel}&\textcircled{1} & \checkmark&73.6/85.7 &69.6/94.0 &70.3/83.9 &75.5/94.6&71.7/89.4\\
          &\textcircled{2}& &71.3/84.2&66.0/91.9&68.0/83.5&74.9/94.7&68.9/87.8\\
          \bottomrule[0.4mm]
    \end{tabular}}
\label{table8}
\end{table*}

\begin{table*}[!t]
\caption{Backbone evaluation on KITTI~\citep{kitti} dataset. \underline{Underline} indicates the default backbone for our P2P-point and P2P-voxel, respectively. \textbf{Bold} denotes the best result.}
\centering
    \resizebox{0.85\linewidth}{!}{
    \normalsize
    \begin{tabular}{c|c|ccccc}
          \toprule[0.4mm]
          \rowcolor{black!10} Tracker& Version& Car & Pedestrian & Van & Cyclist & Mean \\
          \midrule
          \multirow{3}{*}{P2P-point} & \underline{PointNet}~\citep{pointnet}&\textbf{68.8}/\textbf{81.7}&\textbf{62.7}/\textbf{89.1}&\textbf{65.4}/\textbf{80.1}&74.8/94.8&\textbf{66.2}/\textbf{85.4}\\
          &PointNet++~\citep{pointnet++} & 52.7/66.3&22.1/49.8&37.2/44.2&66.3/90.5 &38.4/57.8\\
           &DGCNN~\citep{dgcnn} &68.6/81.1&57.4/85.5&63.8/78.4&\textbf{74.9}/\textbf{94.9}&63.5/83.2 \\
           \midrule
            \multirow{2}{*}{P2P-voxel}&\underline{VoxelNet}~\citep{voxelnet} &\textbf{73.6}/\textbf{85.7} &\textbf{69.6}/\textbf{94.0}& \textbf{70.3}/\textbf{83.9} &\textbf{75.5}/\textbf{94.6}& \textbf{71.7}/\textbf{89.4}\\
            &PointPillars~\citep{pointpillars} & 69.4/82.9&66.1/92.3&69.6/82.1&73.4/94.3 & 68.1/87.2\\
          \bottomrule[0.4mm] 
    \end{tabular}}
\label{table9}
\end{table*}

\noindent\textbf{Influence of Data Augmentation.} To demonstrate the suitability of our data augmentation approach for 3D single object tracking, we re-train P2P-point and P2P-voxel models using the data augmentation introduced in M$^2$Track, \textit{i.e.}, [\textcircled{2}] in Tab.~\ref{table8}. The reported results indicate that our approach is more effective, which can be attributed to two main factors: generating the current search region with the previous target as its center is helpful to simulate the test error; translations along the $xyz$ axes can simulate more fine-grained motion patterns.

\noindent\textbf{Further Evaluation on Different Backbones.} Tab.~\ref{table9} reports tracking results of our P2P-point and P2P-voxel using different backbones. As observed in the upper part, PointNet++~\citep{pointnet++} involves downsampling point clouds, thereby making it difficult to form implicit spatial structure information of point clouds, leading to a significant reduction in tracking performance. In addition, the tracking version using DGCNN~\citep{dgcnn} exhibits a slight performance decrease. This decline may be attributed to the fact that DGCNN extracts local-domain features via graph structures, which may be redundant for pact-to-part motion modeling. In the lower part, PointPillars~\citep{pointpillars} exhibits explicit spatial structure of point clouds, and thus presenting similar pact-to-part motion modeling to P2P-voxel, achieving competitive tracking performance. 

\subsection{Exploration Studies}

\begin{figure}[t]
    \centering
    \includegraphics[width=0.85\linewidth]{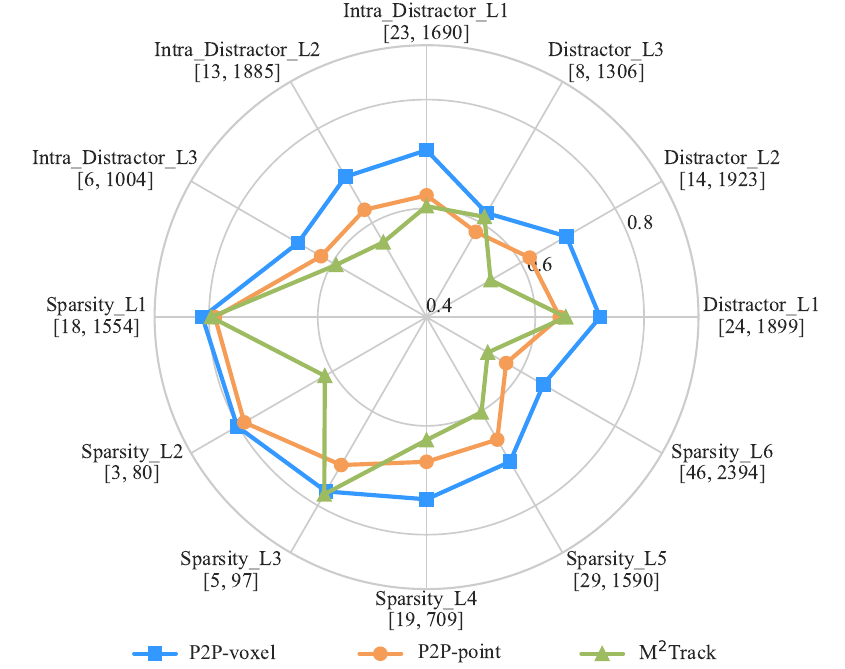}
    \caption{Success rate on sparse and distracting scenes. L$n$ denotes the level of scene complexity. [$m$, $n$] is the number of point cloud sequences and total frames in the corresponding case.}
    \label{fig8}
\end{figure}

\begin{figure}[t]
    \centering
    \includegraphics[width=1.0\linewidth]{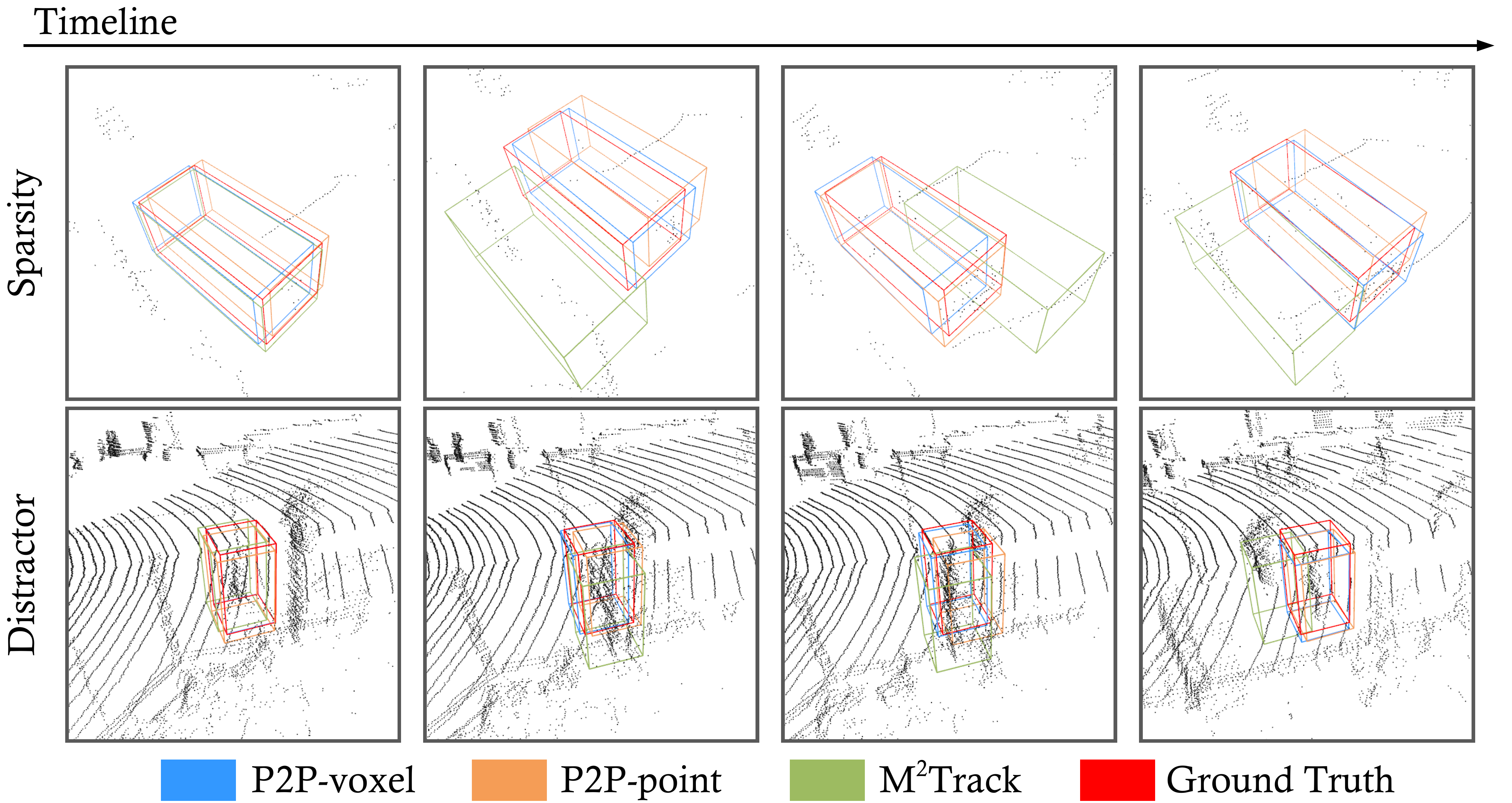}
    \caption{Comparison of tracking results on sparse (upper) and distracting (lower) point cloud scenes, respectively. The proposed P2P-voxel/point can track the targets more accurately and robustly than M$^2$Track~\citep{m2track} in both sparse and distracting tracking scenes.}
    \label{fig9}
\end{figure}

\begin{figure}[t]
    \centering
    \includegraphics[width=0.9\linewidth]{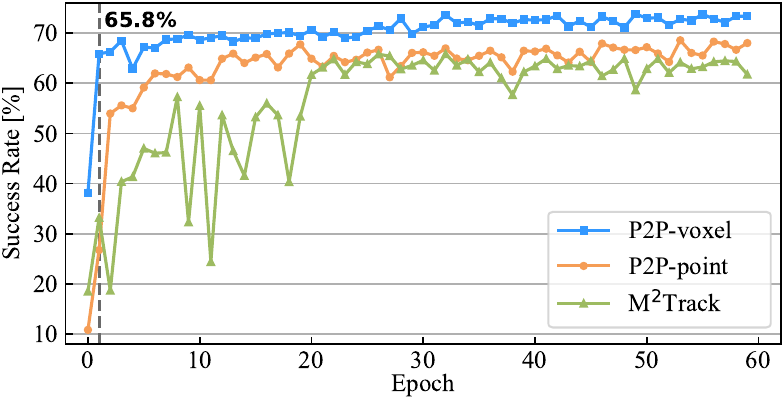}
    \caption{Comparison of model convergence speed.}
    \label{fig10}
\end{figure}

\begin{table*}[t]
\caption{Comparison of tracking results with or without incorporating box information~\citep{m2track} on KITTI~\citep{kitti} dataset. Not including box information is the default setting.}
\centering
    \resizebox{0.8\linewidth}{!}{
    \normalsize
    \begin{tabular}{c|c|c|ccccc}
                \toprule[0.4mm]
          \rowcolor{black!5}Tracker&\# &Box Information & Car & Pedestrian & Van & Cyclist & Mean \\
          \midrule
          \multirow{2}{*}{P2P-point}&\textcircled{1} & &68.8/81.7 &62.7/89.1& 65.4/80.1& 74.8/94.8 & 66.2/85.4\\
          &\textcircled{2} & \checkmark&71.0/83.1&64.2/90.7&68.9/83.4&74.3/94.0&68.0/86.7\\
           \midrule
           \multirow{2}{*}{P2P-voxel}&\textcircled{1} & &73.6/85.7 &69.6/94.0 &70.3/83.9 &75.5/94.6&71.7/89.4\\
          &\textcircled{2}& \checkmark&75.3/86.9&69.7/93.8&72.1/85.0&75.2/94.8&72.7/90.0\\
          \bottomrule[0.4mm]
    \end{tabular}}
\label{table_box}
\end{table*}

\noindent\textbf{Robustness to Sparsity and Distractors.} In real scenarios, LiDAR point clouds are usually sparse and contaminated with distractors. Hence, it is necessary to analyze the tracker's robustness to sparse point clouds and distractors. The Car category data is divided into sequences of varying sparsity levels, following~\citep{p2b,bat,glt}. Likewise, the Pedestrian category data is classified into three levels based on the number of distractors and intra-class distractors, respectively. Fig.~\ref{fig8} shows that P2P-voxel and P2P-point perform better in sparse scenes compared to M$^2$Track, particularly in extremely sparse scenes, such as Sparsity$\_$L4, L5, and L6. Furthermore, our methods demonstrate enhanced resilience to distractors, especially for intra-class distractors. Fig.~\ref{fig9} shows that our P2P-voxel/point can track objects more accurately and robustly, which further intuitively illustrates the superiority of our proposed methods.

\noindent\textbf{Comparison of Convergence Curves.} Fig.~\ref{fig10} illustrates the convergence speed of our proposed methods and M$^2$Track~\citep{m2track} on the KITTI Car category. Our curves jitter more rapidly compared to M$^2$Track, indicating that the proposed models exhibit faster convergence and greater stability. Furthermore, due to P2P-voxel enables explicit part-to-part fusion between the spatial structures of point clouds in consecutive frames, the Success rate reaches 65.8\% in 2 epochs. 

\noindent\textbf{Effectiveness of Box Information.} Box size information is a valuable feature for enhancing tracking performance, as demonstrated in BAT~\citep{bat} and M$^2$Track~\citep{m2track}. Here, we incorporate box information into our proposed P2P-point and P2P-voxel models to evaluate its impact. Following BAT and M$^2$Track, we calculate the pairwise Euclidean distance between each point and the 9 key points of the bounding box. As shown in Tab.~\ref{table_box}, incorporating box information significantly improves tracking performance, particularly for the car and van categories. This phenomenon can be attributed to the robustness and accuracy of box information for grid-like objects compared to non-grid objects. Additionally, we observe a larger performance gain in P2P-point than in P2P-voxel. This is because the voxel representation used in P2P-voxel is inherently more effective for handling rigid objects, and its performance has already reached saturation on the KITTI dataset.

\begin{table*}[t]
\centering
\caption{Comparison on varying levels of sparsity on Car category of KITTI~\citep{kitti} and NuScenes~\citep{Nuscenes}.  \textbf{Bold} denotes the best result.}
    \resizebox{0.8\linewidth}{!}{
    \normalsize
    \begin{tabular}{c|cccccc}
          \toprule[0.4mm]
          \rowcolor{black!10} Dataset & \multicolumn{6}{c}{KITTI Car}  \\
          \midrule
          \rowcolor{black!10} Interval & [0, 10) & [10, 20) & [20, 30) & [30, 40) & [40, 50) & [50, +$\infty$) \\
          \midrule
         \rowcolor{black!10} Sequence Number & 46  & 29  & 19 & 5  & 3 & 18 \\
          \midrule
          \rowcolor{black!10}  Frame Number & 2,394 & 1,590 &  709 &  97 & 80 & 1,554 \\
          \midrule
          M$^2$Track~\citep{m2track}&  53.0/67.1 & 60.2/73.9& 62.6/75.7 & \textbf{77.6}/\textbf{92.1}&61.6/72.1 & 79.6/92.1\\
         P2P-point & 56.9/68.9&66.0/79.0&66.6/82.0&71.4/87.3&78.7/91.1 & 78.9/91.0  \\
           P2P-voxel &\textbf{64.8}/\textbf{75.0}&\textbf{70.7}/\textbf{83.6}&\textbf{73.5}/\textbf{87.8}&77.0/90.4&\textbf{80.2}/\textbf{92.7} & \textbf{81.2}/\textbf{92.7} \\
           \midrule
           \midrule
            \rowcolor{black!10} Dataset & \multicolumn{6}{c}{NuScenes Car} \\
          \midrule
          \rowcolor{black!10} Interval &  [0, 10) & [10, 20) & [20, 30) & [30, 40) & [40, 50) & [50, +$\infty$)\\
          \midrule
         \rowcolor{black!10} Sequence Number& 2,884 & 212  & 101  & 64  & 37  & 362\\
          \midrule
          \rowcolor{black!10}  Frame Number &  45,322& 4,375 &  2,190 & 1,321 & 814 &  10,127\\
          \midrule
          M$^2$Track~\citep{m2track} & 52.1/60.9&57.1/65.9 & \textbf{65.8}/\textbf{73.4} &  68.1/76.3& 70.1/78.6 & \textbf{75.7}/\textbf{83.1} \\
         P2P-point & 59.0/65.3&62.9/70.1 & 63.9/68.9& \textbf{70.4}/\textbf{78.2}& 71.8/78.2 & 73.5/80.0 \\
           P2P-voxel & \textbf{63.3}/\textbf{71.4}&\textbf{64.1}/\textbf{71.9} & 64.3/70.2&69.6/77.4 & \textbf{73.9}/\textbf{79.9}& 72.8/79.4\\
          \bottomrule[0.4mm] 
    \end{tabular}}
\label{table10}
\end{table*}

\begin{figure*}[t]
\begin{center}
\includegraphics[width=0.4\linewidth]{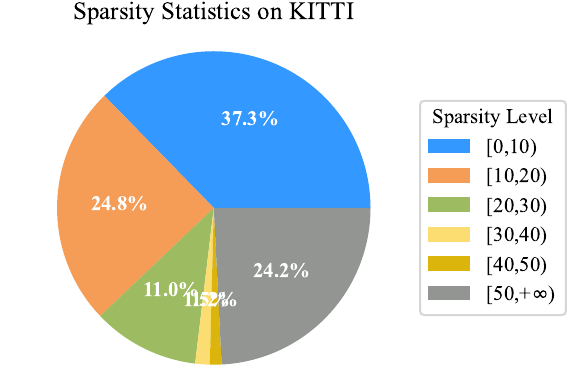} \quad
\includegraphics[width=0.4\linewidth]{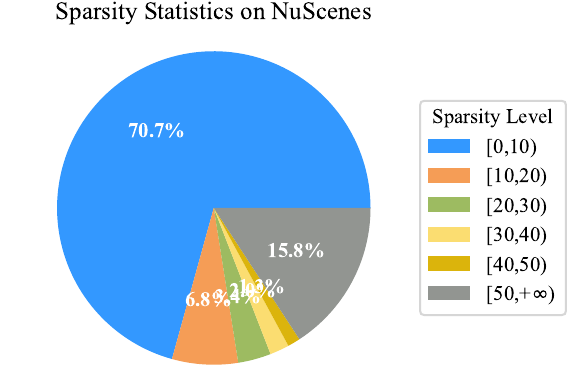}
\end{center}
\caption{Sparsity statistics in in the \textbf{Car category} of KITTI~\citep{kitti} and NuScenes~\citep{Nuscenes} datasets. A point cloud sequence is categorized as sparse if the first frame contains less than 50 points. We categorize sparsity into five distinct levels.}
\label{fig11}
\end{figure*}

\begin{figure}[t]
  \centering
   \includegraphics[width=0.95\linewidth]{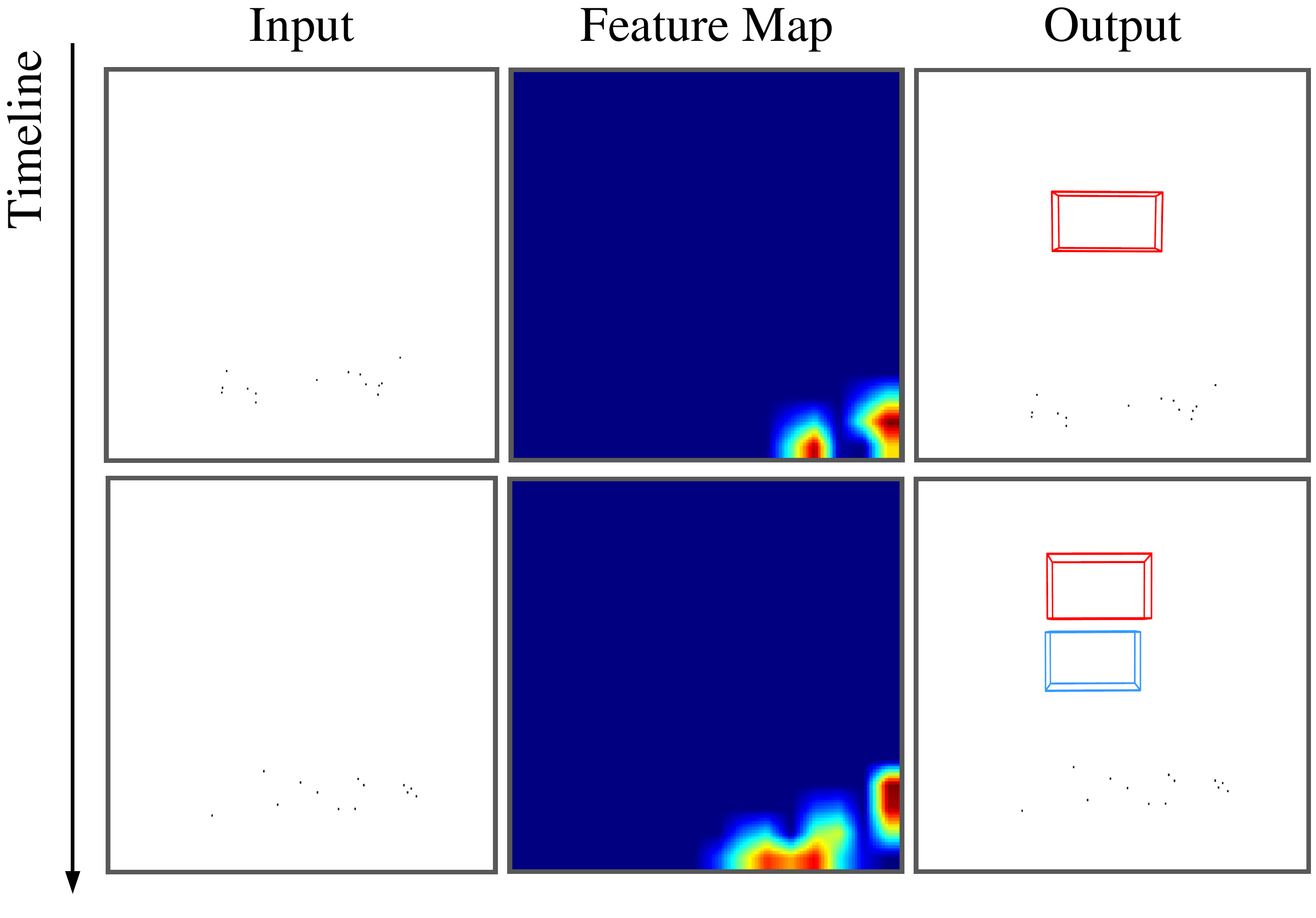}
   \caption{An instance of failure in the proposed P2P-voxel. We plot input point cloud, feature map on $XY$ plane and output tracking box. The The \textbf{\textcolor{lightblue}{blue box}} and \textbf{\textcolor{red}{red box}} represent our predicted result and ground truth, respectively.}
   \label{fig12}
\end{figure}

\section{Limitation and Discussion}
As illustrated in Tab.~\ref{table2} and~\ref{table3}, tracking methods seem to have nearly reached a state of performance saturation on the KITTI~\citep{kitti} dataset. Nevertheless, on the NuScenes~\citep{Nuscenes} dataset, there is still significant room for developing tracker. Our findings suggest that this phenomenon is largely influenced by the different sparsity in the two datasets. We count the sparsity distribution in the Car category of both the KITTI and NuScenes datasets, as shown in Fig.~\ref{fig11}. Moreover, tracking performance corresponding to different sparsity levels is reported in Tab.~\ref{table10}. 

We have the following three observations: First, in the case of sparse scenes with fewer than 40 points, there is a notable performance decrease in the tracking algorithms; Second, our proposed models demonstrate better robustness to sparse scenes compared to M$^2$Track~\citep{m2track}, especially in extremely sparse scenarios, such as [0,10) and [10,20); Third, NuScenes contains significantly more sparse scenes compared to KITTI. Therefore, accurately tracking objects in sparse scenes remains a challenging issue.

\noindent\textbf{Limitation.}
The proposed P2P framework has significantly advanced 3D single object tracking, outperforming previous state-of-the-art methods by a significant margin. Despite this progress, challenges remain evident. Fig.~\ref{fig12} illustrates a tracking failure of our P2P-voxel method. Tracking drift occurs when target information becomes obscured due to occlusion or extended distance scanning by the LiDAR sensor. Tab.~\ref{table10} also reflects this phenomenon. This inspires two considerations for future tracking development:
\begin{itemize}
    \item \textbf{Model Considerations}: Introducing temporal motion information could mitigate information loss. However, this solution depends on several frames with relatively accurate motion priors. Alternatively, exploring multimodal tracking, where 2D image data complements the loss of 3D point cloud information, may offer a better solution. 
    \item \textbf{Dataset Considerations}: Dataset-centric solutions are also vital. Creating higher-quality datasets by more advanced LiDAR sensors is a promising avenue for further investigation.
\end{itemize}

\section{Conclusion}
This paper introduces P2P, a strong framework for 3D SOT. The novel framework regards 3D single object tracking problem as a direct inference of target relative motion between consecutive frames. It performs part-to-part fusion between spatial structures of consecutive point clouds to explore fine-grained information changes across the corresponding parts, modeling accurate motion cues. Extensive experiments demonstrate our P2P framework is efficient and achieves superior performance over the previous state-of-the-art trackers. We expect this work could serve as a baseline framework for future work.


\vspace{2ex}
\noindent\textbf{Acknowledgment} 
This work was partially supported by the Institute of Information \& communications Technology Planning \& Evaluation (IITP) grant funded by the Korea government(MSIT)(RS-2020-II201373, Artificial Intelligence Graduate School Program (Hanyang University)), and the Zhejiang Provincial Key Laboratory of Intelligent Vehicle Electronics Research.

\bibliography{sn-bibliography} 

\end{document}